\documentclass[10pt,journal,compsoc]{IEEEtran}
\renewcommand\IEEEkeywordsname{Keywords}
\usepackage[utf8]{inputenc}
\usepackage{graphicx}
\usepackage{tikz}
\usepackage{array}
\usetikzlibrary{trees,positioning,shapes,shadows,arrows}
\usepackage{cite}
\usepackage{amsmath,amssymb,amsfonts}
\usepackage{pifont}
\usepackage{bbding}
\newcommand{\cmark}{\text{\ding{51}}}
\newcommand{\xmark}{\text{\ding{55}}}
\DeclareMathOperator*{\argmin}{arg\,min}
\usepackage{algorithmic}
\usepackage{graphicx}
\usepackage{subcaption}
\usepackage{textcomp}
\usepackage{booktabs}
\usepackage{multirow}
\usepackage{multicol}
\usepackage{makecell}
\usepackage{color}
\usepackage{float}
\usepackage{url}

\usepackage{hyperref}
\usepackage{cite}
\usepackage{latexsym}
\usepackage{ragged2e}

\begin{document}

\title{Predictive Modeling of Hospital Readmission: Challenges and Solutions\\
}

\author{Shuwen Wang,
        Xingquan Zhu,~\IEEEmembership{Senior Member,~IEEE}
        \thanks{Preprint version accepted by IEEE/ACM Trans. on Computational Biology and Bioinformatics (TCBB)\protect}
}




\IEEEtitleabstractindextext{%
\justify
\begin{abstract}
Hospital readmission prediction is a study to learn models from historical medical data to predict probability of a patient returning to hospital in a certain period, \textit{e.g.} 30 or 90 days, after the discharge. The motivation is to help health providers deliver better treatment and post-discharge strategies, lower the hospital readmission rate, and eventually reduce the medical costs. Due to inherent complexity of diseases and healthcare ecosystems, modeling hospital readmission is facing many challenges. By now, a variety of methods have been developed, but existing literature fails to deliver a complete picture to answer some fundamental questions, such as what are the main challenges and solutions in modeling hospital readmission; what are typical features/models used for readmission prediction; how to achieve meaningful and transparent predictions for decision making; and what are possible conflicts when deploying predictive approaches for real-world usages. In this paper, we systematically review computational models for hospital readmission prediction, and propose a taxonomy of challenges featuring four main categories: (1) data variety and complexity; (2) data imbalance, locality and privacy; (3) model interpretability; and (4) model implementation. The review summarizes methods in each category, and highlights technical solutions proposed to address the challenges. In addition, a review of datasets and resources available for hospital readmission modeling also provides firsthand materials to support researchers and practitioners to design new approaches for effective and efficient hospital readmission prediction. 
\end{abstract}
\IEEEoverridecommandlockouts

}

\maketitle
\IEEEdisplaynontitleabstractindextext

\IEEEpeerreviewmaketitle

\ifCLASSOPTIONcompsoc
\IEEEraisesectionheading{\section{Introduction}\label{sec:introduction}}
\else
\section{Introduction}
\label{sec:introduction}
\fi

\IEEEPARstart{H}ospital readmission is defined as a hospital visit of a discharged patient being admitted again to the same or a different medical institution within a specific period of time, such as 30 days or 90 days, following the previous visit. A revisit usually implies an incomplete or unsuccessful treatment from the previous in-patient visit, therefore it is a defined metrics of the US healthcare system~\cite{healthrankingwebsite}. 
Reasons behind hospital readmissions are complicated and various, among which many are avoidable especially those related to doctors, nurses, and healthcare system. For example, patients discharged ahead of the schedule are more likely to be readmitted~\cite{holly2016}. Medication errors, like no proper prescriptions for necessary medicines when a patient was discharged, are also responsible for preventable readmissions\cite{foster2003,foster2004}. In addition, socioeconomic factors, including patients of both lower and higher socioeconomic status, are tied to the readmission risk\cite{wier2006}, and a disparity in hospital readmission between ethnicity has also been observed \cite{tsai2014}, in which the possibility that Medicare patients, being readmitted to hospital after major surgeries, among Black patients is 19\% higher than White patients. 

A succession of hospital revisits, in a short period, are costly to patients and healthcare system\cite{healthrankingwebsite}. 
Avoidable hospital readmission not only incurs \$41.3 billion annual cost\cite{hcupwebsite}, it also places signature pressure to the medical resources as well as the high working intensity of medical staff. 

\begin{table*}[h]
    \small
    \caption{A summary of predictive models used for hospital readmission prediction.}
    \centering
    \scriptsize
    \begin{tabular}{p{0.15\textwidth}|p{0.26\textwidth}|p{0.25\textwidth}|p{0.25\textwidth}}
    \hline
    \hline
    {Predictive Model Types} & Methods \& Papers&Strength&Weakness\\
    \hline
    Clinical rule based methods&{LACE Index, HOSPITAL score, B Score} \cite{sd2017paper24,kh2016paper37,jd2016paper40,ll2017paper48,mh2017paper49,jd2013hospitalscore,ss2016paper54,mj2017paper55,sy2015paper60,pc2012paper61,xm2019paper67,gg2013paper74,rb2017paper89,er2016paper90,lp2014paper97,dm2019Bscore,Elixhauser1998ComorbidityMF,charlson1987score,hq2005coding,cw2009modify}&
    High transparency \& interpretability&Low accuracy \& limited discriminative power\\
    \hline
    Case-based reasoning &{$k$-NN classifiers} \cite{cb2017cost}&
    High interpretability. Easy to maintain \& adapt to changes & Time-consuming with high-dimensional medical data\\ 
    \hline        
    Regression based methods&{Logistic Regress} \cite{tsai2014,sa2013,v2012,e2013,kz2013,mjpaper14,gg2016comparing,jp2016paper16,cv2010lace,ch2014paper20,ey2016paper22,sd2017paper24, gs2017,dk2014paper26,br2008paper27,lg2014paper30,oh2010paper32,cf2009paper34,ck2018paper36,kh2016paper37,mt2001paper43,mb2015paper47,ll2017paper48,ke2014paper50,ms2013paper52,kz2015paper53,ss2016paper54,sy2015paper60,pc2012paper61,MM2015paper63,rm2015paper64,ep1999paper65,xm2019paper67,jf2016paper70,jh2005paper72,gg2013paper74,pc2011paper76,ms2008paper80,mk2012paper81,eg2013paper83,jm2013paper85,cb2013paper92,sw2009paper96,lp2014paper97,ma2018supervised,hz2018different,dh2013mining,JF1997paper56,db2020flow,lb2018gender3,ll2018gender1,rr2019race2,bk2005race3}&
    Easy to interpret \& efficient to train with large patient records&Ineffective for modeling nonlinear relationships\\
    \hline    
    Decision tree methods&{Decision tree, Classification and regression tree (CART)} \cite{a2013,kh2016paper37,kz2015paper53,ss2016paper54,sf2015paper71,nf2020paper75,jh2014paper95,ma2018supervised,db2020flow}&
    Transparency \& Interpretability. Automatic identifying important factors&Unstable \& sensitive to data. Low accuracy\\
    \hline
    Bayesian methods&{Naive bayes, Bayesian conditional probability} \cite{kh2016paper37,vm2015,pw2019,a2013, pw2019,ad2019rhythms,rd2016,ks2017,MM2015paper63,ma2018supervised,yl2017tf1}&
    Transparency \& incorporate domain knowledge&Computationally expensive \& low accuracy\\ 
    \hline            
    Neural networks&{multi-layer neural networks, RBF network} \cite{dk2014paper26,mj2017paper55,rd2016,kz2013paper62,pw2019,mm2008training,ma2018supervised,wl2020code,ec2016medical,ppc2020sr2}& 
    High accuracy \& effective for high dimensional data&Low training efficiency. Poor transparency \& interpretability\\    
    \hline    
    Margin classifiers \& kernel machines &{Support vector machines (SVM)} \cite{jf2015comparison, lt2016paper3,sjj2018,ac2018paper23,jz2013paper31,ss2016paper54,sy2015paper60,xm2019paper67,arbj2019machine,ma2018supervised,tb2018federated}&
    Model complex high-dimensional patient records&Low interpretability for clinical decisions\\
    \hline        
    Ensemble methods&{Bagging, Boosting, Random forest, Gradient boosting} \cite{vb2015,kh2016paper37,ss2016paper54,xm2019paper67,ad2019rhythms,jf2015comparison,sjj2018,kz2013,ac2018paper23,ss2016paper54,xm2019paper67,sb2016UsingPC,ac2018paper23,arbj2019machine,ky2020predicting,db2020flow}&
    Improve accuracy over single models&High computational costs \& lack of transparency due to combined decisions\\
    \hline        
    Cost-sensitive classification&{Bayesian optimal decision} \cite{cb2017cost,hw2018cost}&
    Tackle imbalanced class distributions \& consider medical costs in modeling&Time-consuming on enormous skewed data\\
    \hline        
    NLP methods &{Topic models} \cite{ar2016nlp}&
    Work with unstructured data like discharge summaries&Cannot handle structured data\\
    \hline            
    Deep learning methods&{CNN, LSTM, deep contextual embedding} \cite{cx2018,bk2018,sg2018,hw2018cost,arbj2019machine,sa2020embpred30,xz2017localized,ar2018patientembed,zh2019vectors}&
    High accuracy. Modeling complex relationships across multiple visits& Long training time for parameter tuning \& require large volumes of data\\
    \hline            
    Clustering methods&{Hierarchical clustering} \cite{rm2015underlying,lh2019patient,co2016clustering,sh2015paper19,jj2018sr1}&
    Do not require labels. Interpret data distributions for large volume data &Ineffective for high dimensional data\\
    \hline                
    \hline
    \end{tabular}
    \label{tab:table 1}
\end{table*}

In order to mitigate the severity of high readmission rates, the US federal government introduced a series of plans\cite{planswebsite}. Combining payment and readmissions through Hospital Readmissions Reduction Program (HRRP) is part of the initiatives to reduce readmission rates\cite{hrrpswebsite}. 

Following the HRRP initiative, many researches have been conducted to tackle the problem. One commonly used approach is to apply predictive models, such as logistic regression, to learn from historical data, and then predict the readmission possibility of a patient after being discharged from hospital\cite{e2013,sa2013,v2012}. Other advanced machine learning methods, such as decision trees\cite{a2013} and deep learning\cite{bk2018}, are also used for readmission prediction. Table \ref{tab:table 1} summarizes commonly used predictive models in the field. In general, eleven popular predictive model types are considered such as clinical rule based method, case-based reasoning, regression based method and deep learning methods. Among all those methods, majority research uses regression based methods (logistic regression), neural networks, and ensemble methods including bagging, boosting, random forest and gradient boosting. Due to complications of human diseases, some models are developed based on specific disease types, like heart failure\cite{vb2015,kz2013,sg2018}, pneumonia\cite{gs2017}, and organ transplantation\cite{mm2013}.

Indeed, many predictive models haven been proposed and are reportedly effective under certain circumstances, but they often do not perform well as expected when being applied to new health records\cite{dk2011,bg2017}. Evidently, Medicare, under the HRRP plan, cut payments to 2,853 hospitals in 2019. Among 3,129 general hospitals being evaluated in the HRRP program, 83\% of them received a penalty. Partially, this is because that a hospital readmission is a compounded outcome of many factors, and not all of them can be modeled by using computational approaches. On the other hand, most existing methods only focus on the modeling and learning aspects of the problem, failing to address the underlying challenges for readmission prediction. A recent study~\cite{em2020use} systematically reviews 41 readmission prediction models (including 17 models for all patient risk prediction and 24 models for patient specific populations). Their investigations suggest that using Electronic Medical Records (EMR) data have better predictive performance than those using administrative data. However, technical challenges and solutions of readmission prediction still remain unaddressed, and are largely unclear.

The above observations motivate our study to review challenges and solutions for hospital readmission prediction. In order to systematically address the challenges, we propose a taxonomy to categorize selected modeling methods, and summarize how existing approaches handle different challenges. In addition to the review of methodologies, we also outline public datasets available for model building and evaluation. The survey provides a comprehensive review 
for researchers and practitioners to understand the state-of-the-art in the filed, as well as designing new approaches to tackle hospital readmission prediction. 


The rest of the paper is organized as follows. Sec. \uppercase\expandafter{\romannumeral2} proposes a taxonomy of hospital readmission challenge. Secs. \uppercase\expandafter{\romannumeral3} to \uppercase\expandafter{\romannumeral6} outline each challenge as well as methods used to tackle them, including algorithms, applications, and performance. Public datasets are summarized in Sec. \uppercase\expandafter{\romannumeral7}, and we conclude the paper in Sec. \uppercase\expandafter{\romannumeral8}.

\begin{figure*}[h]
\begin{small}
    \centering
    \includegraphics[width=185mm,height=45mm]{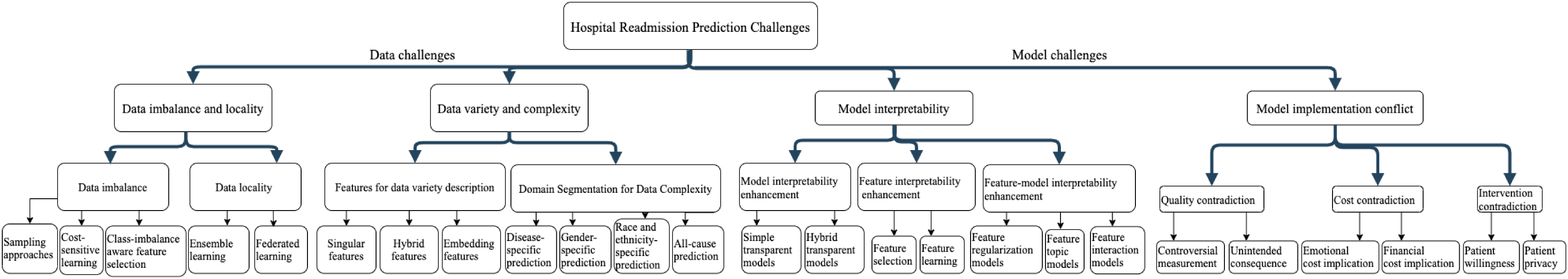}
    \caption{A taxonomy of challenges of computational methods for hospital readmission prediction. }
    \label{fig:taxonomy}
\end{small}
\end{figure*}

\section{Problem Definition \& Taxonomy}
\subsection{Problem Definition}
Formally, we use an $m$ dimensional vector $\mathbf{x}_i^{[a_j,d_j]}\in\mathbb{R}^m$ to denote a patient $i$ and his/her $j^{th}$ hospital visit, where $[a_j,d_j]$ denote admission and discharge/disease time of the $j^{th}$ visit, respectively. The $m$ dimensional vector includes symptoms, treatment, medical notes, medications, procedures, and a variety of electronic health record (EHR) information of the patient, carried out during the $j^{th}$ visit. A \textit{hospital readmission} of a patient $i$ is referred to a visit $j$, whose admission time $a_j$ is within a certain window, typically 30 days or 90 days, following the discharge of the previous visit $d_{j-1}$, \textit{i.e.} $a_j-d_{j-1}\le 30$. Given a number of patients and their visit records, hospital readmission prediction \textbf{aims} to accurately predict the readmission probability of a patient after being discharged from the current visit.

\subsection{Taxonomy of Readmission Prediction Challenges}
Accurate prediction of hospital readmission is a significant challenge, mainly because health records and diseases are inherently complex in nature. For example, not all medical records are organized in feature format, and medical treatments and procedures are complicated making feature engineering a daunting task. Meanwhile, HIPPA (Health Insurance Portability and Accountability Act) regulations and policies also raise challenges for data usage and sharing. 
In order to carry out systematic review of computational models for hospital readmission prediction, we propose a taxonomy in Fig. \ref{fig:taxonomy} to summarize main technical challenges into two types and four categories: data challenges 
and model challenges. 
Data challenges explain the nature of readmission prediction and data related issues for model learning. Model challenges, on the other hand, explain model training, interpretation, and implementation issues. 
\subsubsection{Data Challenges}
Data imbalance, locality and privacy represent the first data challenge for hospital readmission prediction. Databases used for hospital readmission prediction usually contain a large number of patient visit records collected from healthcare providers across disease types, ages, as well as length of in-patient treatment \cite{nrdswebsite,niswebsite}. Although patients' medical records are gathered and presented in details, the class distributions are often imbalanced\cite{sjj2018} meaning readmission visits are much fewer than normal visits. Such large-scale but imbalanced data is an obstacle for learning accurate predictive models\cite{znl2019}. In addition to the data imbalance, the regional designation of hospitals adds additional complexity to this issue. Demographics of the hospital served regions often have great disparity across regions, where the top two characteristics reflecting the difference are race and average income of the residents. In other words, due to regional restrictions, some hospitals may have majority patients from certain ethnic groups, like Latino or Asian. Such difference greatly prompts the locality of hospital, and further complicates the data challenge. Medical information collected from those hospitals is biased for readmission research. Besides, the income gap between residents also affects medical expenses\cite{ck2013,energencyswebsite}. Low income cohort may be vulnerable to medical services such as medical insurance, implying that there may exit a higher possibility that they are not willing to continue their medical treatment after discharge, even their conditions do not improve. This resistance to treatment makes the collected data lack integrity, and impose challenges to predictive modeling. 

Data variety and complexity is another challenge to predict readmission. In order to train predictive models, it is necessary to collect information/features to characterize object of interest (such as patient, disease, hospitals, patient visits \textit{etc.}). Due to inherent complexity of the healthcare systems, and dependency between readmission and other items, such as diseases, comorbidities, preexisting conditions, \textit{etc.}, a predictive model should consider a variety of data objects for learning\cite{sg2018}. In addition to the data variety, a hospital readmission is a compounded outcome of many issues, including patient life styles, disease types, hospital treatments \textit{etc.}, making accurate readmission prediction a complex task. Accordingly, data resources and learning may choose to focus on different medical conditions, by using domain segmentation, such as disease-specific prediction, gender-specific prediction \textit{ect.}. 

\subsubsection{Model Challenges}
At the model level, transparency and conflicts must be properly resolved before predictive models can be put into real-world usages. 
Interpretability of a model refers to comprehension of model decision process for human understanding\cite{tm2019}. A model resembling to human perception and decision logic, such as a decision rule or tree, is always preferred in medical domains. 
From practical aspect, it is also important to understand why some patients are predicted as high risk after being discharged from the hospital while others are not. 
Many models perform well on benchmark datasets, but their decision logistic is intransparent. As a result, their performance may deteriorate after being applied to new data, due to the blind decision process.

The second model challenge is to resolve conflicts between models and implementation, such as qualify of services, cost reductions, patient emotion, \textit{etc.}. The HRRP initiative intends to encourage hospitals to reduce their readmission rates\cite{jw2017}. It is expected that, as unnecessary readmission being reduced, the net income of hospitals can increase significantly\cite{su2019}. However, the reduction of hospital readmission does not directly imply better qualify of services. A short-term cost reduction may result in a higher cost in the long run, and jeopardy the HRRP objective. 
Meanwhile, as more stakeholders are turning to predictive models, patients data are being collected for sharing and analysis\cite{ic2014}, resulting in increasing privacy concerns\cite{45cr2012}. 

\section{Data Imbalance, Locality and Privacy}
Data imbalance and locality are two common biases in medical data, 
which are known to impose significant challenge to predictive models~\cite{mm2008training}.  

\subsection{Data Imbalance}
Data imbalance refers to a phenomenon where datasets used to train a predictive model have a biased class distribution. In many cases, one type of samples (\textit{i.e.} positive class) are significantly less than other types of samples. This is partially caused by the reality that disease samples are only a small percentage of the whole population, and naturally results in the class imbalance. Learning models with imbalanced class distributions is defined challenge, because most algorithms are affected by frequency bias and pay more attention to majority class samples~\cite{hh2009learningfrom}.  
Data imbalance tends to force the classifier to classify all samples as normal, in order to satisfy the defined objective function, such as minimizing the classification errors\cite{pb2015asurveyofpreedictive}. Common solutions are to rebalance samples in different classes, by manipulating data populations (sampling approaches) or classification outcomes (cost-sensitive learning). 

\subsubsection{Sampling Approaches}
Sampling approaches change data distributions to balance samples in different groups in order to tackle the data imbalance challenge. Common sampling solutions are to either drop majority class samples, repeat samples from minority class, or create synthetic samples for minority class.

\begin{figure*}[h]
\begin{small}
\begin{subfigure}{.19\textwidth}
    \centering
    \includegraphics[width=0.8\linewidth]{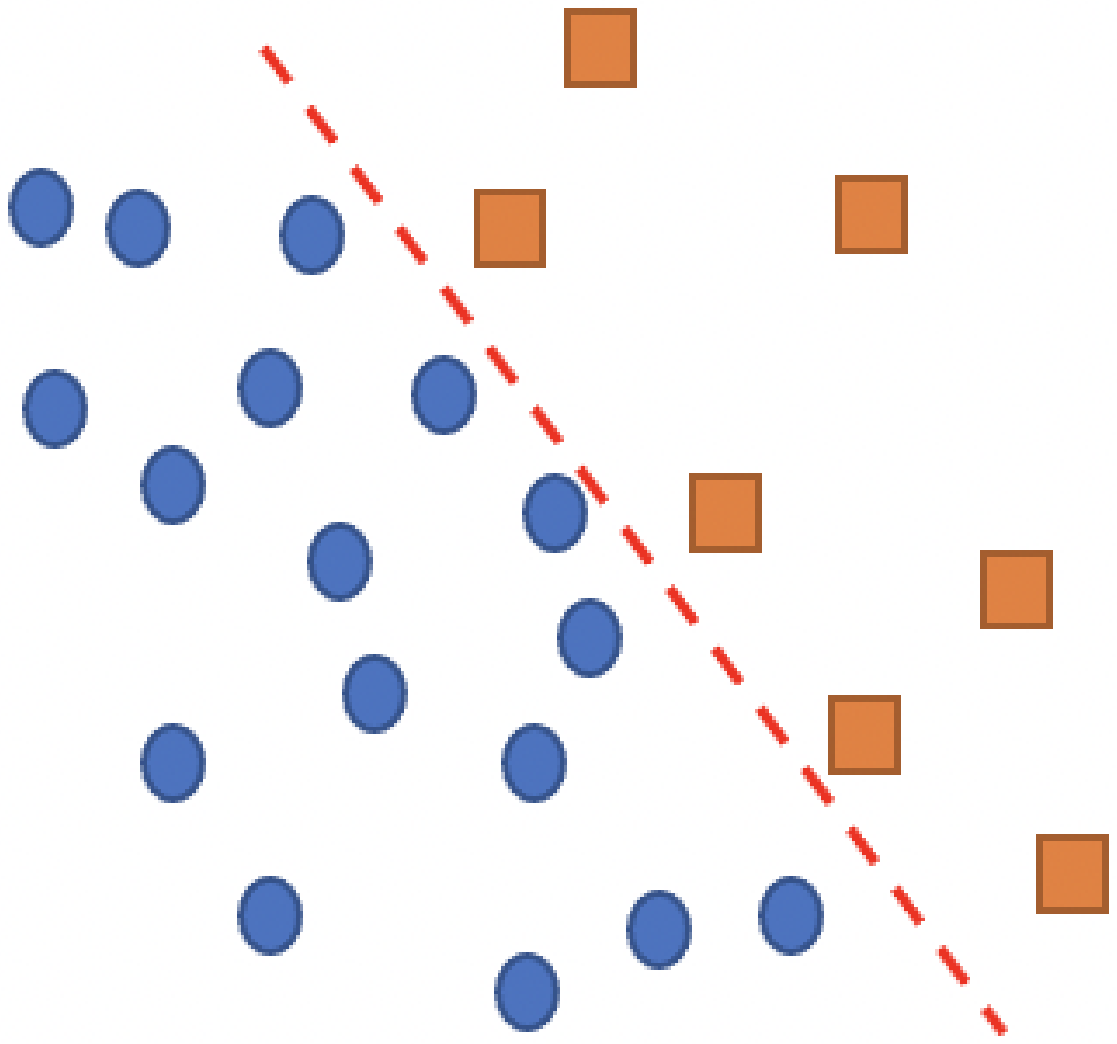}
    \caption{Original dataset}
    \label{fig:(a)}
\end{subfigure}
\begin{subfigure}{.19\textwidth}
    \centering
    \includegraphics[width=0.8\linewidth]{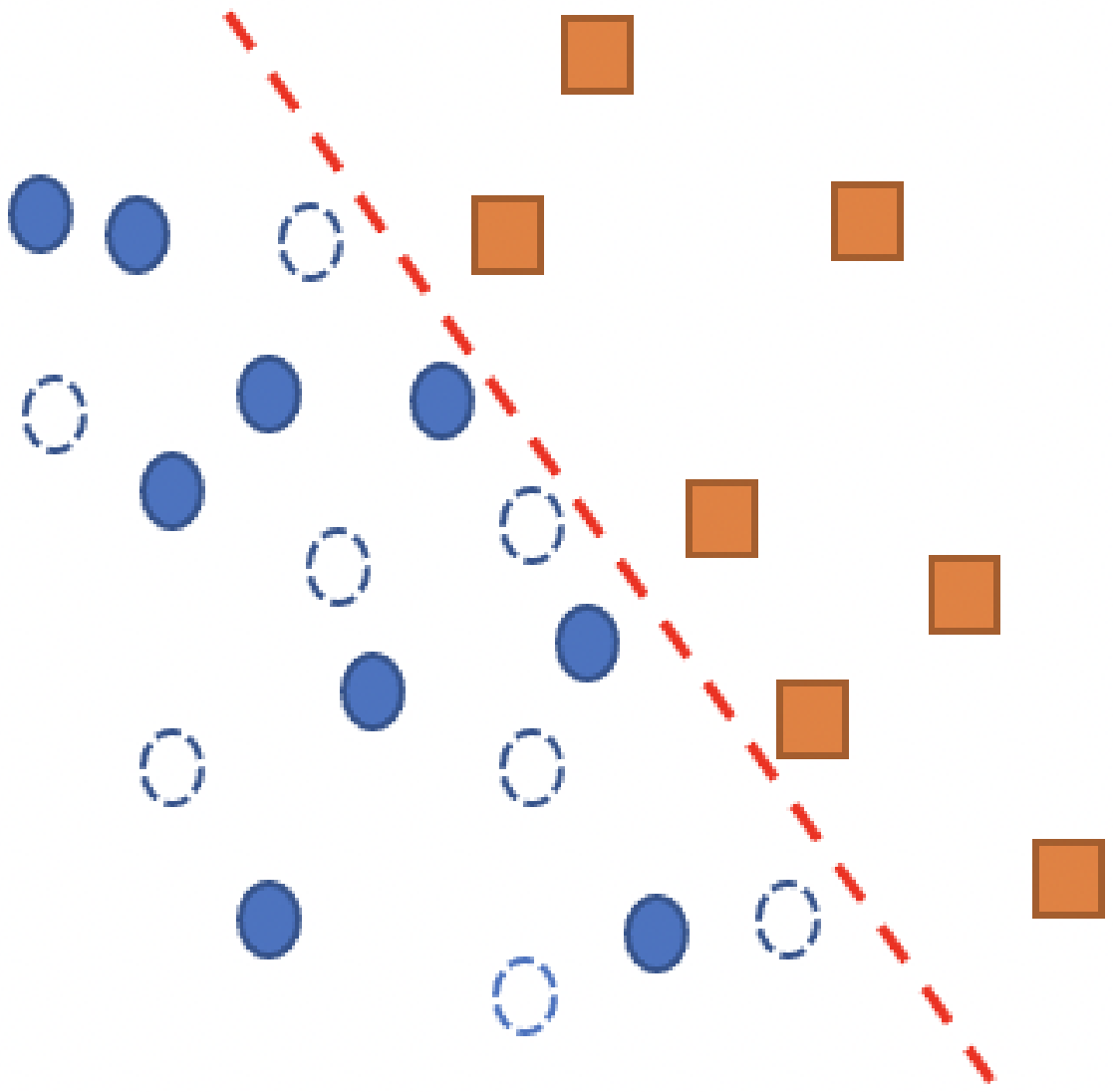}
    \caption{RUS sampling}
    \label{fig:(b)}
\end{subfigure}
\begin{subfigure}{.19\textwidth}
    \centering
    \includegraphics[width=0.8\linewidth]{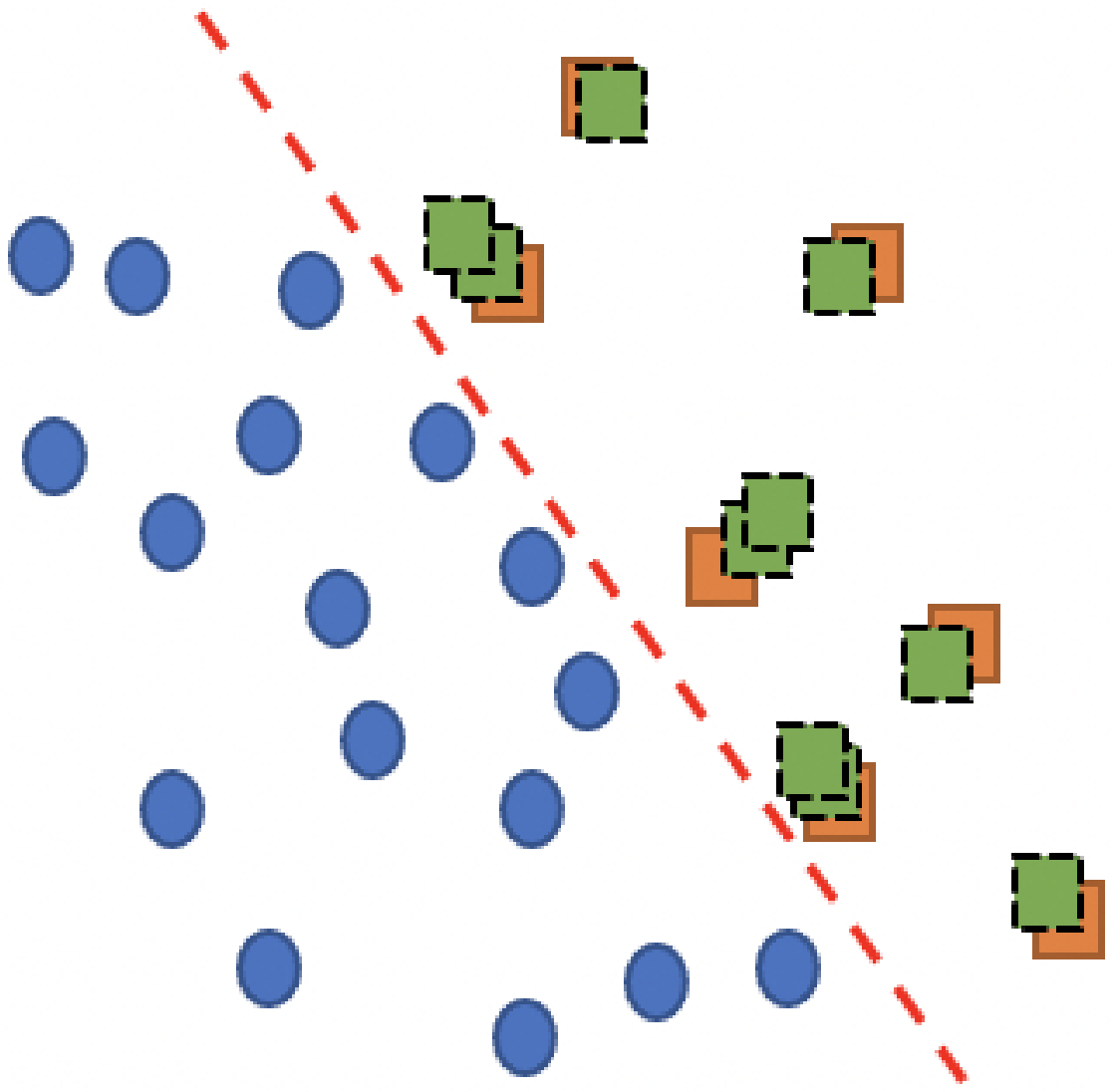}
    \caption{ROS sampling}
    \label{fig:(c)}
\end{subfigure}
\begin{subfigure}{.19\textwidth}
    \centering
    \includegraphics[width=0.9\linewidth]{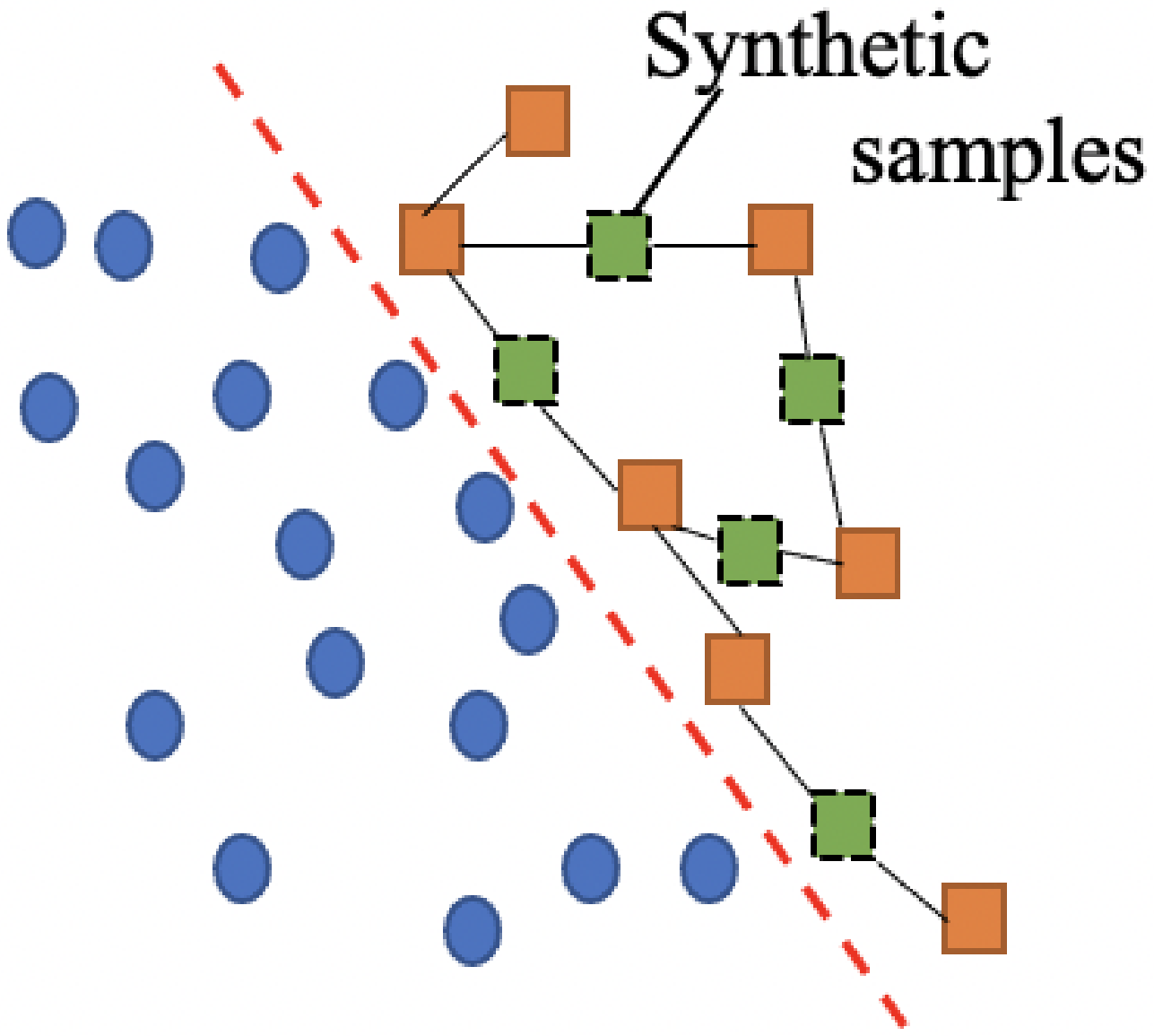}
    \caption{SMOTE}
    \label{fig:(d)}
\end{subfigure}
\begin{subfigure}{.2\textwidth}
    \centering
    \includegraphics[width=0.9\linewidth]{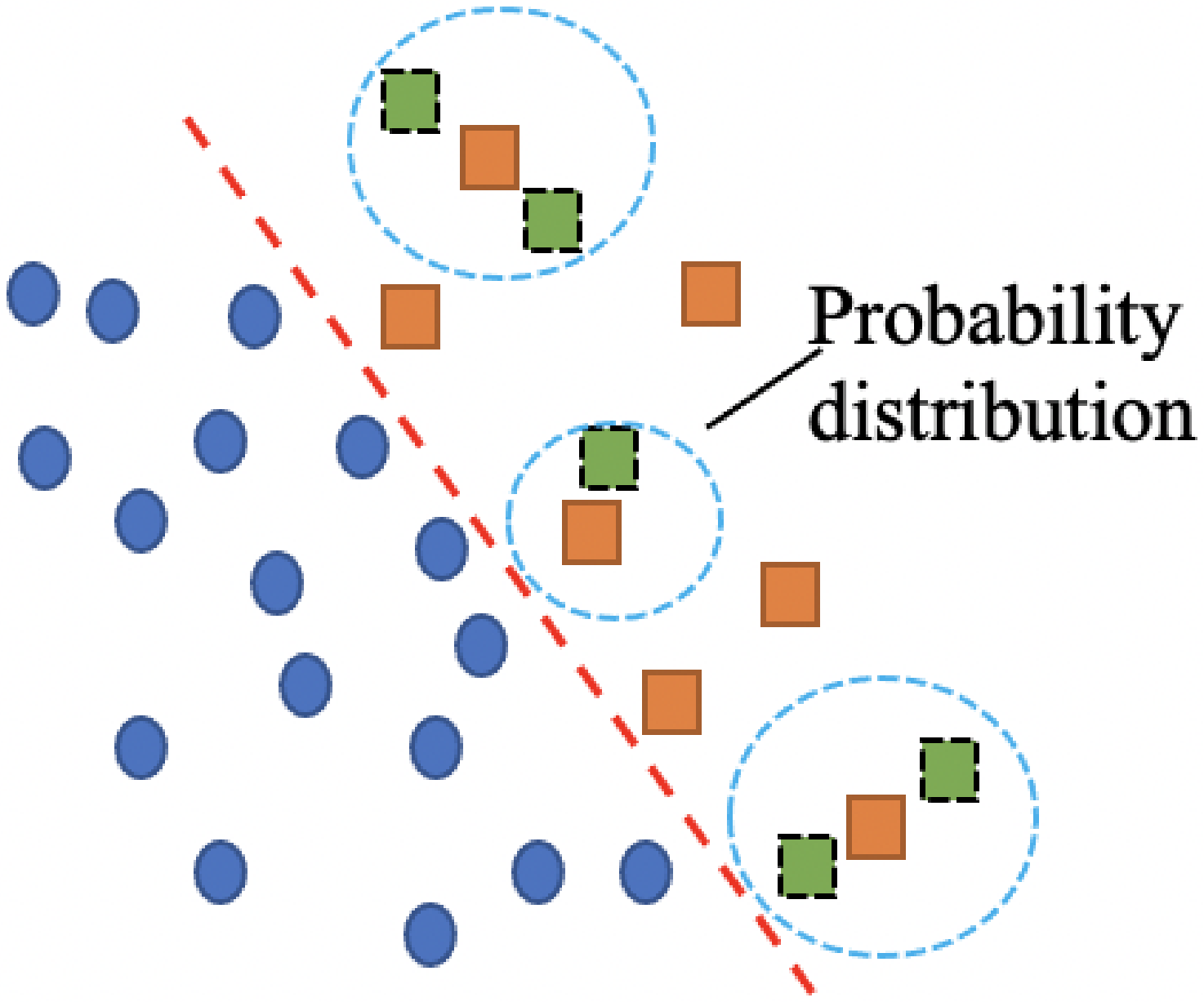}
    \caption{ROSE}
    \label{fig:(e)}
\end{subfigure}
\caption{(a) A dataset with imbalanced class distributions; (b) random under sampling (RUS); (c) random oversampling (ROS); (d)Synthetic Minority Over-sampling Technique (SMOTE); and (e) Random Over Sampling Examples (ROSE).}
\label{FIGURE sampling}
\end{small}
\end{figure*}

\textbullet \underline{Random Sampling Approaches}
Random Oversampling (ROS) and Undersampling (RUS) are the simplest ways broadly used in numerous domains, such as hospital readmission prediction and fraud detection. Fig. \ref{FIGURE sampling} describes the sampling methods in which Fig. \ref{fig:(a)} represents the original imbalanced data with blue dots as the prevalent class and orange rectangular as the minority class. Random undersampling, in Fig. \ref{fig:(b)}, involves randomly selecting examples from the majority class and down sampling them in the training dataset according to sampling strategy defined as the ratio of the minority class to majority class. Due to the loss of vast quantities of discarded data, a loss in classification performance can be resulted from the boundary ambiguity between the two classes. Random oversampling, in Fig. \ref{fig:(c)}, duplicates minority samples to form a balanced training set. Due to sample duplication, ROS may lead to over-fitting in the training.

A research study~\cite{arbj2019machine} using Medical Information Mart for Intensive Care III (MIMIC-III) database~\cite{mimic} shows that, by using undersampling, their model achieves 0.642 AUC score for ICU patient readmission. Another study~\cite{ma2018supervised} investigates RUS sampling and five supervised learning methods, decision trees, naive bayes, logistic regression, neural networks, and support vector machines (SVM) for risk modality and hospital readmission prediction. The results show that, overall, neural networks achieve best performance for both risk modality and hospital readmission prediction. In addition, using AdaBoost to change the weight of instances for learning results in 3\% and 6\% improvement for readmission and mortality predictions, respectively.

\textbullet \underline{Synthetic Samples}
For RUS and ROS sampling, the dropped/duplicated instances are part of the original training data, meaning that there is a potential risk that sampling will introduce information loss or bias. Synthetic sample generation, on the other hand, will generate new samples similar (but not identical) to the training data. Two common approaches to generate synthetic samples are Synthetic Minority Over-sampling Technique (SMOTE)~\cite{nc2002smote} and Random Over Sampling Examples (ROSE)~\cite{mg2014training}. As shown in Figs. \ref{fig:(d)} and \ref{fig:(e)}, SMOTE is an improvement of Random over sampling approach. A minority class sample $x$ is randomly selected, with its $k$ nearest minority class neighbors being determined. Then the synthetic instance is created by choosing one of the $k$ nearest neighbors $b$ at random and connecting $x$ and $b$ to form a line segment in the feature space. The synthetic instances can be generated as a convex combination of the instances $x$ and $b$. ROSE, on the other hand, generates synthetic samples using a conditional density estimating the positive and negative classes. A randomly selected minority sample is used as the center of a created density function, and synthetic samples are the ones generated from the estimated density functions.

Using SMOTE to generate synthetic instances to balance positive and negative samples for 30 day readmission prediction has been studied\cite{sa2020embpred30} by using a UCI hospital readmission dataset~\cite{uci130}. The experiments show exceptionally higher AUC values (0.974) than results from other studies (normally around 0.7 AUC range). One possible reason is that UCI readmission dataset has a relatively balanced sample distributions because 11.2\% samples belong to positive class (readmission), whereas in other dataset, such as National Readmission Database~\cite{nrdswebsite}, the positive ratio is much smaller. By using different sampling approaches, including RUS, ROS, and ROSE, a method~\cite{ac2018paper23} comparatively studies the three methods using UCI readmission dataset~\cite{uci130}, using different classifiers, such as SVM, random forest, gradient boosting, and regression and partition trees. The results 
show that ROSE is significantly worse than other approaches (including original data without any sampling). In addition, RUS and ROS have comparable performance, and both frequently outperform models trained from original imbalanced dataset.

\textbullet \underline{Random Class Balancing}
A new technique to tackle data imbalance called Random Balance is proposed inspired by the idea of randomly deciding class proportions \cite{dp2015balance}. In this approach, Data sampled from training dataset is used to train every member of the Random Balance ensemble and the augmentation is completed with synthetic samples created by SMOTE. The probabilities of selecting an instance from minority class and majority class are presented in Eq.~(\ref{eq: min balance} ), 
where $N$ is the total samples with $p$ positive instances and $n$ negative instances.
\begin{equation}
   P_{mi} = \frac{1}{N-3}(N-\frac{p+3}{2}-\frac{1}{p}); P_{ma} = \frac{1}{N-3}(N-\frac{n+3}{2}-\frac{1}{n})
    \label{eq: min balance}
\end{equation}


Although preprocessing techniques are usually used to restore the balance of the class proportions to a given level, Random Balance relies on completely random proportions. The key step in this method is that both the size of majority and minority classes is randomly set, followed by SMOTE and Random Undersampling in order to increase or reduce the size to match the ideal class size. This methods alleviates the problem of deleting important examples by being repeated multiple times. Despite its simplicity, this methods outperforms other advanced ensemble methods.

\subsubsection{Cost-Sensitive Learning}
Data imbalance challenge can also be mitigated through the change of learning algorithms. The ultimate goal of machine learning is to minimize/satisfy the loss function. If the misjudgment loss on minority samples is increased in the loss function, the model can be adjusted to better identify minority samples. Cost sensitive learning is one approach to adjust the loss function of the learning algorithm to make the model sensitive to minority samples.

Cost-sensitive confusion matrix, in Table \ref{tab:table 2}, is a common way to adjust the loss function. Instead of treating all misclassification equally, the cost matrix differentiates costs associated to different types of mistakes. For example, in Table \ref{tab:table 2}, $\mu $ is defined as the cost of a single False Positive (FP) and $\lambda$ is defined as the cost of a single False Negative (FN). In order to support cost-sensitive readmission prediction, one can either apply the cost matrix to the posterior probability (classification outcomes) or the learning objective function to minimize the misclassification costs.  

\begin{table}[h]
    \caption{Cost sensitive confusion matirx. $\lambda$ and $\mu$ are positive values denoting misclassification costs between two groups (positive \textit{vs.} negative).}
    \centering
    \scriptsize
    \begin{tabular}{c|c|c}
    \hline
                    & Predicted positive & Predicted negative \\
\hline
Actual positive & 0 & $\lambda$  \\
\hline
Actual negative &  $\mu$  & 0 \\   
    \hline
    \end{tabular}
    \label{tab:table 2}
\end{table}

\textbullet \underline{Posterior Probability Adjustment}: In a research \cite{cb2017cost} predicting 30-day readmission for patients with Chronic Obstructive Pulmonary Disease (COPD), the cost matrix is used to adjust the posterior provability for prediction. Assuming $x$ denotes a patient (or a visit) and $P(True|x)$ denotes the posterior probability of a generic classifier classifying $x$ as being a readmission (True), the prediction of readmission is based on the adjustment of posterior probability in Eq.~(\ref{eq:costclassi}) where the threshold of classification is presented as $\frac{\mu}{\lambda+\mu}$. 
\begin{equation}
  \textit{Prediction} = \left\{
  \begin{array}{cc}
    \text{True},& \text{If}~ P(True|x) > \frac{\mu}{\lambda+\mu} \\
    \text{False},& \text{Otherwise}
  \end{array}\right.
  \label{eq:costclassi}
\end{equation}
By applying the above design to several generic classification models, including Naïve Bayes (NB), Random Forest (RF), Support Vector Machines (SVM), $k$-Nearest Neighbors (kNN), C4.5, Bagging with REPTree, and Boosting with Decision Stump, their experiments\cite{cb2017cost} show that cost-sensitive classification is effective in minimizing costs and the cost matrix is more desirable than commonly used AUC, when evaluating hospital readmission systems.

\textbullet \underline{Learning Objective Function Adjustment}: In order to directly integrate the costs to the learning, a cost-sensitive formulation\cite{hw2018cost} is used to train a multi-layer perceptron fed by learnt features through convolutional neural networks (CNN) and statistical features via feature embedding to predict hospital readmission. To tackle the misclassification problems of the minority class, the cost sensitive deep neural network (CSDNN) consists of one input layer, one output layer, and multiple hidden layers with fully-connected neurons formulated using weight matrix $\mathbf{W}$. A modified cross entropy is set as the loss function shown in Eq.~(\ref{eq:cost}), where $n$ is the total number of patients, $y_i\in{\{T,F\}}$ denotes the label of a patient $x_i$, where True (1) means a readmission or False (0) otherwise.  $P(\hat{y_i}|\mathbf{x}_i,\mathbf{W})$ denotes the posterior probability of the $i^{th}$ patient, and $\mathbf{C}(\hat{y_i},y_i)$ defines the classification cost of $x_i$ with respect to the current prediction $y_i$.
\begin{equation}
    \mathcal{L} = -\frac{1}{n}\sum_{i=1}^n \log[\sum_{\hat{y_i}\in{\{T,F\}}}P(\hat{y_i}\vert\mathbf{x}_i,\mathbf{W})\mathbf{C}(\hat{y_i},y_i)]+\frac{\lambda}{2}\Vert\mathbf{W}\Vert_2^2
    \label{eq:cost}
\end{equation}
The above approach is validated on two real-world medical datasets from Barnes-Jewish Hospital and the results prove that their prediction of readmission performs significantly better than several baselines with a much higher Area Under the ROC Curve (0.70 AUC score) than baselines.

\subsubsection{Class-imbalance aware feature selection approach}
A readmission prediction algorithm Joint Imbalanced Classification and Feature Selection (JICFS) is proposed to construct the loss function and applied sample weight to handle class-imbalance problem \cite{Du2020JointIC}. To be specific, this approach solves the readmission prediction problem with imbalance class by constructing an improved margin-based loss function, which involves two parameters $\alpha$ and $\gamma$ to reduce the weight of loss assigned to easily classified samples.
\begin{equation}
    \min_{w\in\mathbb{R}^m}f(w) =  \sum_{n=1}^N(\alpha_nlog(1+exp(-y_n\gamma(w^Tz_n)))/\gamma)+\lambda\Vert{w}\Vert_1
    \label{eq: imbalance aware}
\end{equation}

In Eq.~(\ref{eq: imbalance aware}), the class-imbalance aware feature selection approach contains $\alpha$, $\gamma$ and $\lambda$ three parameters from objective function to obtain coefficient matrix $w$ in order to realize feature selection from class-imbalance data. The method was compared with different class-imbalance learning algorithms based on six real-world readmission datasets and it always can achieve better performance on each dataset.

\subsection{Data Locality and Privacy}
While data imbalance is concerning the learning target (or class labels), data locality, on the other hand, is associated to the sample distributions (or independent variables). At population level, data for readmission prediction might be collected from a local/regional hospital, where the demographics of the patient body naturally introduce bias. 
At individual level, when collecting data for each patient, the hospital visits used to characterize the patient may also introduce bias. At the ministration level, regularizations also impose restriction for data sharing across hospitals, making it difficult to learn good models from local data.  

A study~\cite{hz2018different} considers two types of discharge sampling, first time discharge \textit{vs.} all discharges. In other words, the research compares using a single discharge \textit{vs.} using all discharges of patient visits to represent each patient for learning. Experiments show that using the first discharge per patient underestimates the readmission rate, and may result in misleading measures of model performance.

Common approaches to tackle data locality and privacy are to employ ensemble learning or federated learning. The former trains multiple models from local datasets and combine them for prediction, whereas the latter trains one model from multiple decentralized/localized datasets.

\subsubsection{Ensemble Learning}
Ensemble learning 
combines multiple base models for prediction. Typical approaches include bagging, boosting, and stacking~\cite{do1999popular}. Bagging trains base models separately (often in parallel), and then combines them using weighted (or unweighted) majority voting. Boosting, on the other hand, trains base models in a sequential and progressive manner, so a later trained base model is improved based on an earlier trained base model. Stacking is a meta learning approach, which uses base classifiers to generate outputs, and then retrains another model from the outputs for prediction.

In~\cite{xz2017localized}, a localized sampling approach is proposed to allow sampling process to focus on instances difficult to classify. By using localized sampling to generate balanced datasets, this approach is validated using data collected from several South Florida regional hospitals. A joint ensemble-learning model~\cite{ky2020predicting} combines weight boosting algorithm with stacking algorithm, and compares three major baseline (1) the LACE index, (2) RandomForest-Lasso-SMOTE, and (3) SMOTE (which uses SMOTE to replace bagging for data samping) on national Hosptial Quality Monitoring System (HQMS) database (including 651,816 records after data processing). The results show that LACE (which is commonly hospital score systems) has the least performance, confirming that machine learning is useful for hospital readmission prediction. Meanwhile, bagging with weight boosting and stacking shows clear benefits on high dimensional medical data with imbalance class distributions and imbalanced misclassification costs.  

\subsubsection{Federated Learning} Different from ensemble learning which focuses on combining models trained from local datasets, federated learning tackles data locality challenge by allowing multiple data holders to collaboratively train a model, and keep participant data in private without exchanging raw data. A research~\cite{tb2018federated} proposes to use federated learning to build a global model to predict hospitalizations due to heart diseases using patient electronic health records (whether a patient will be hospitalized within one or two years, prior to the time of prediction). 
To tackle the problem, they formulate the problem as a sparse support Vector Machine (SVM) learning problem, with the following objective function:
\begin{equation}
    \min_{\textbf{w},w_0}\sum_{i=1}^n f_i(\textbf{w},w_0,x_i)+0.5\tau\left\Vert \textbf{w} \right\Vert^2_2+\rho\left\Vert w_0 \right\Vert_1
    \label{eq:ssvm}
\end{equation}
In Eq.~(\ref{eq:ssvm}), $\textbf{w}\in\mathbb{R}^d$ and $w_0 \in\mathbb{R}$ are weight vectors (parameters) defining the classifier. $f_i(\textbf{w},w_0,x_i)=\max\{0,1-\ell(\textbf{w}^T x_i+w_0)\}$ defines a hinge loss for instance $x_i$. $\tau$ and $\rho$ are penalty coefficients enforce L$_2$ norm and L$_1$ norm constraints on the parameters. In federated learning setting, multiple agents (or hospitals) each hold their own private data, so $x_i,i=1,\cdots,n$ are not presented to any single agent. The key is to use local data from each single agent, to learn $(\textbf{w},w_0)$, which optimizes the objective function in Eq.~(\ref{eq:ssvm}). To solve their own $(\textbf{w},w_0)$, based on local data, all agents will combine learned solutions to create a global $(\textbf{w},w_0)$. In addition to the predictive model, their method also find important factors/features associated to hospitalizations, such as ``Age'' and ``Admission due to Other Circulatory System Diagnoses'' etc.

Similarly, another research~\cite{lh2019patient} studies federated learning for patient mortality and hospital stay time prediction, using distributed electronic medical records. They propose a distributed clustering to separate patients into clinically meaningful communities (both communities and data are local). Experiments show that this approach results in a higher predictive accuracy and lower communication cost, comparing to other federated learning methods. 

\section{Data variety and Complexity}
A hospital readmission is the outcome of numerous compounding factors, involving patients, diseases, care providers \textit{etc.}. Collecting representative training data is an important step for modeling. Therefore, the second major data challenge for readmission prediction is to properly characterize training samples and learning tasks.

\subsection{Features for Data Variety Description}
Patient records from electronic medical records (EMRs) and administrative databases usually include various information from basic personal demographics to professional medical diagnosis\cite{kt2014datasets}. 
In this section we summarize features used to describe samples into three major categories: singular features, hybrid features, and latent features. The detailed feature types are listed in Table \ref{tab:table 3}, including demographics, admission and discharge, clinical, hospital \textit{etc.} 

\begin{table*}[h]
    \centering
    \small
    \caption{A summary of features used to describe patients, hospitals, hospital visits for readmission prediction.}
    \scriptsize
    \begin{tabular}{p{0.29\textwidth}|p{0.2\textwidth}|p{0.4\textwidth}|p{0.05\textwidth}}
    \hline
    \hline
    Feature Types \& Reference & Feature Subtypes & Feature Information & Structured\\
    \hline
    \multirow{7}{*}{\parbox{0.2\textwidth}{Demographics features\newline \noindent\cite{jf2015comparison,sg2018,lt2016paper3,sa2013,v2012,e2013,vb2015,sjj2018,a2013,kz2013,gg2016comparing,jp2016paper16,cv2010lace,bb2012paper18,sh2015paper19,ch2014paper20,ey2016paper22,ac2018paper23,sd2017paper24,gs2017,dk2014paper26,br2008paper27,sm2015paper28,mm2013,jz2013paper31,oh2010paper32,rd2015paper33,cf2009paper34,ck2018paper36,kh2016paper37,vm2015,jd2016paper40,pw2019,mt2001paper43,dh2013mining,rk2004paper46,mb2015paper47,ll2017paper48,ke2014paper50,jd2013hospitalscore,ms2013paper52,kz2015paper53,ss2016paper54,mj2017paper55,JF1997paper56,bk2018,sy2015paper60,pc2012paper61,kz2013paper62,MM2015paper63,rm2015paper64,ep1999paper65,ks2017,xm2019paper67,rd2016,jf2016paper70,sf2015paper71,jh2005paper72,gg2013paper74,nf2020paper75,pc2011paper76,dm2015paper79,ms2008paper80,mk2012paper81,eg2013paper83,jm2013paper85,fk2015paper86,rb2017paper89,er2016paper90,cb2013paper92,sb2016UsingPC,jh2014paper95,sw2009paper96,lp2014paper97,ph2006paper98,lb2018gender3,ll2018gender1,rd2015gender2,rr2019race2,bk2005race3}}}&\multirow{2}{*}{Basic demographics information}&Age, Gender, Race, Education, Marital status&\cmark\\
    \cline{3-4}
    &&Language, Income/Financially issues&\cmark\\
    \cline{2-4}
    & \multirow{1}{*} {Contact information}&Family size/members, Zip code/address, Family doctor&\cmark\\
    \cline{2-4}
    & \multirow{1}{*} {Insurance information}&Insurance provider, Mode of payment&\cmark\\
    \cline{2-4}
    & \multirow{2}{*} {General health information}&Medical conditions, Allergies, Current medications&\xmark\\
    \cline{3-4}
    &&Completed outpatient appointment rate, Nursing home needed&\cmark\\
    \cline{2-4}
    & \multirow{1}{*} {Social history}&Smoking, Alcohol, Living situation, Employment&\xmark\\
    \hline
    \multirow{4}{*}{\parbox{0.3\textwidth}{Admission and discharge information\newline
    \noindent\cite{jf2015comparison,sg2018,lt2016paper3,sa2013,v2012,e2013,vb2015,sjj2018,a2013,kz2013,gg2016comparing,jp2016paper16,cv2010lace,bb2012paper18,sh2015paper19,ch2014paper20,ey2016paper22,ac2018paper23,sd2017paper24,gs2017,dk2014paper26,br2008paper27,sm2015paper28,mm2013,lg2014paper30,jz2013paper31,oh2010paper32,rd2015paper33,cf2009paper34,ck2018paper36,kh2016paper37,vm2015,jd2016paper40,pw2019,mt2001paper43,dh2013mining,rk2004paper46,mb2015paper47,ll2017paper48,mh2017paper49,jd2013hospitalscore,ms2013paper52,kz2015paper53,ss2016paper54,mj2017paper55,JF1997paper56,bk2018,sy2015paper60,pc2012paper61,kz2013paper62,MM2015paper63,rm2015paper64,ep1999paper65,ks2017,xm2019paper67,rd2016,jf2016paper70,sf2015paper71,jh2005paper72,gg2013paper74,pc2011paper76,cx2018,dm2015paper79,ms2008paper80,mk2012paper81,eg2013paper83,jm2013paper85,fk2015paper86,rb2017paper89,er2016paper90,cb2013paper92,sb2016UsingPC,jh2014paper95,sw2009paper96,lp2014paper97,ph2006paper98,lb2018gender3,ll2018gender1,rd2015gender2,rr2019race2,bk2005race3}}}&\multirow{2}{*}{Admission information}&Admission date, First hospital visit, Elective&\cmark\\
    \cline{3-4}
    &&Number of admissions in a past time period, Cost-weight of previous admission, Diagnosis of last admission&\cmark\\ 
    \cline{2-4}
    & \multirow{2}{*} {Discharge information}&Total charge, Discharge date&\cmark\\
    \cline{3-4}
    &&Transfer, Discharge disposition&\cmark\\
    \hline
  \multirow{5}{*}{\parbox{0.3\textwidth}{Clinical information\newline
  \noindent\cite{ad2019rhythms,sb2016UsingPC,sb2016UsingPC,jf2015comparison,sg2018,lt2016paper3,sa2013,v2012,e2013,vb2015,sjj2018,a2013,mjpaper14,gg2016comparing,jp2016paper16,cv2010lace,sh2015paper19,ch2014paper20,sr2015paper21,ey2016paper22,ac2018paper23,sg2018,dk2014paper26,br2008paper27,mm2013,lg2014paper30,jz2013paper31,oh2010paper32,rd2015paper33,cf2009paper34,ck2018paper36,kh2016paper37,vm2015,jd2016paper40,pw2019,mt2001paper43,dh2013mining,rk2004paper46,mb2015paper47,ll2017paper48,ke2014paper50,jd2013hospitalscore,ms2013paper52,kz2015paper53,ss2016paper54,JF1997paper56,sy2015paper60,pc2012paper61,kz2013paper62,MM2015paper63,rm2015paper64,ep1999paper65,ks2017,xm2019paper67,rd2016,jf2016paper70,sf2015paper71,jh2005paper72,gg2013paper74,nf2020paper75,pc2011paper76,dm2015paper79,ms2008paper80,mk2012paper81,eg2013paper83,jm2013paper85,fk2015paper86,rb2017paper89,cb2013paper92,rm2015underlying,jh2014paper95,sw2009paper96,lp2014paper97,ph2006paper98,lb2018gender3,ll2018gender1,rd2015gender2,rr2019race2,bk2005race3}}}&\multirow{2}{*}{Payment code information}&ICD-10codes, ICD-9 codes&\cmark\\ 
  \cline{3-4}
  &&APR-DRG codes, DRG codes&\cmark\\
    \cline{2-4}
    &In-hospital symptom&Vitals and lab values&\cmark\\
    \cline{2-4}
    &{Rhythmic features}&Mean 10 most active hours, Total sleep time, Sedentary time&\cmark\\
    \cline{2-4}
    &Medical images&Ultrasound exam&\xmark\\
    \hline
    \multirow{3}{*}{\parbox{0.3\textwidth}{Hospital information\newline
    \noindent\cite{jf2015comparison,gg2016comparing,pw2019,dh2013mining,ep1999paper65,rm2015underlying,bk2005race3}}}&\multirow{2}{*}{Hospital statistics}&Total number of admissions, Number of patients&\cmark\\
    \cline{3-4}
    &&Percent readmission within a time period (30 days etc.)&\cmark\\
    \cline{2-4}
    &Hospital characteristics&Ownership of hospital, Rural/Urban&\cmark\\
    \hline
    Textual information\newline
    \noindent\cite{cx2018,sg2018}&\multicolumn{2}{p{0.6\textwidth}|}{Discharge summary, Physician note, Date the note was initiated, Subject of the note, Prescription medication, Date of the prescription update, Dose and strength of the drug}&\multirow{2}{*}{\xmark}\\
    \hline
    Hybrid information\newline
    \noindent\cite{sd2017paper24,kh2016paper37,jd2016paper40,ll2017paper48,mh2017paper49,jd2013hospitalscore,ss2016paper54,mj2017paper55,sy2015paper60,pc2012paper61,xm2019paper67,gg2013paper74,rb2017paper89,er2016paper90,lp2014paper97,dm2019Bscore,Elixhauser1998ComorbidityMF,charlson1987score,hq2005coding,cw2009modify,rr2019race2}&\multicolumn{2}{p{0.6\textwidth}|}{LACE, HOSPITAL score, Charlson comorbidity score, Elixhauser Comorbidity Index, Baltimore Score}&\cmark\\
    \hline
    Latent features:\cite{dk2014paper26,ar2018patientembed}&\multicolumn{2}{p{0.6\textwidth}|}{Clinical nodes embedding features, Patient embedding features}&\cmark\\
    \hline
    \hline
    \end{tabular}
    \label{tab:table 3}
\end{table*}

\subsubsection{Singular Features}
Singular features include the first five types of features listed in Table \ref{tab:table 3}. They are single factors, indicators, statistics or textual information, used to describes patients, diseases, medical procedures, hospitals \textit{etc.}. Demographics is defined as features including basic patient information such as age, race, contact information in case emergency condition happens, insurance information indicating whether a patient has insurance or not. Such demographics provides general health information representing patients' health condition at the time of the hospital visit. 

After a patient is admitted to hospital, admission and discharge information will record administrative features related to the visit, such as the dates of admission/discharge and whether the patient is admitted through emergent department and so on. Clinical information records patients' symptoms and procedures for in-patient treatment. The information of the hospitals to which patients are admitted are summarized as hospital information, which includes statistic features such as percent readmission within a time period, hospital locations, and ownership. 

Textual information is another type of singular features consisting of literal summary like discharge summary, physician notes. Such information is often stored in unstructured format, and provide comprehensive information not detailed in administrative and clinical features, which are often stored in structured format (such as a table). For most methods, they simply combine single features into one feature set to train predictive models. 

Singular features are ideal for training logistic regression classifiers because features are rather independent and easy to interpret.  A research\cite{kz2015paper53} uses conditional logistic regression, combined with patient demographics, clinical information and categorical data, for model development. After correcting the data imbalance using undersampling, the applicability of conditional logistic regression is tested and compared with several standard classifiers, including standard logistic regression, random forests, SVM and stepwise logistic regression. The models are performed on two different prediction variable sets, original variables and selected ones, in order to achieve efficient readmission management with identified features.

Another similar study\cite{jf2016paper70} also employs demographics, clinical data, and textual notes from administrative data to predict 30-day hospital readmission for maintenance hemodialysis patients. Data are collected from University of North Carolina Hospital (2008 to 2013) and are recorded as means and SDs for continuous variables whereas textual variables are presented as frequencies and percentages. The study trains models using univariate and multivariable binary logistic regressions which only consider variables with a univariate P value \textless0.20. Results show that two multivariable logistic regression models outperform  univarible models with an AUC of 0.79 (95\% CI, 0.73 to 0.85).

\subsubsection{Hybrid Features}
Apart from features representing single factors, hybrid features have a long history being used in the medical field to combine multiple factors to form a feature indicator for prediction. LACE index\cite{cv2010lace}, HOSPITAL score\cite{jd2013hospitalscore}, Charlson comorbidity score\cite{charlson1987score}, Elixhauser  Comorbidity  Index\cite{Elixhauser1998ComorbidityMF} and Baltimore  Score (B Score)\cite{dm2019Bscore}, are the commonly used hybrid features. In many predictive models, hybrid features are further integrated as independent variables to train classifiers for prediction.

\textbullet \underline{Charlson Comorbidity Index}\cite{charlson1987score} was first proposed in 1987 to classify prognostic comorbidity in longitudinal study. In the original version, 19 types of comorbidities are identified and assigned weights according to the adjusted relative risk of one-year mortality. The final single comorbidity score is the sum of weights. ICD-9-CM codes are assigned for categories of the Charlson Comorbidity Index and the number of categories is reduced from 19 to 17 by combining 'Leukemia' and 'Lymphomas' into 'Any tumor'\cite{rd1992adapting}. Later on, a multi-step process\cite{hq2005coding} is conducted to develop ICD-10 codes to define the index and Table \ref{tab:table 4} shows the revised version of Charlson Comorbidity Index. 

A research \cite{e2013} combines Charlson Comorbidity Index with patients demographics as indicators for readmission risk prediction. A retrospective multiple regression analysis of 958 non-pregnant adults is conducted based on the data abstracted from hospital administrative sources and electronic medical records. After comparing with other models, using AUC values for validation, the study concludes that poly-pharmacy and higher Charlson score at admission can predict readmission risk as good as or better than published risk prediction models.

\begin{table}[h]
    \centering
    \small
    \caption{Latest version of Charlson comorbidity index}
    \scriptsize
    \begin{tabular}{c|c}
    \hline
    \hline
    Score&Comorbidity\\
    \hline
    \multirow{4}{*}{1}&Previous myocardial infarction\\&Cerebrovascular disease\\&Peripheral vascular disease\\&Diabetes without complications\\
    \hline
    \multirow{5}{*}{2}&Congestive heart failure\\&Diabetes with end organ damage\\&Chronic pulmonary disease\\&Mild liver or renal disease\\&Any tumor (including lymphoma or leukemia)\\
    \hline
    \multirow{1}{*}{3}&Dementia; Connective tissue disease\\
    \hline
    \multirow{2}{*}{4}&HIV infection\\&Moderate or severe liver or renal disease\\
    \hline
    6&Metastatic solid tumor\\
    \hline
    \hline
    \end{tabular}
    \label{tab:table 4}
\end{table}

\textbullet \underline{LACE Index}
In order to create an easy-to-use index to quantify patients' risk of readmission or death, after being discharged from hospital, a secondary analysis of a multicentre prospective cohort study is performed based on 48 patient-level and admission-level variables\cite{cv2010lace}. Multivariable logistic regression, fractional polynomial function and points system\cite{ls2004system} are applied to derive and validate the LACE Index in Table \ref{tab:table 5} which is the sum of the scores from four aspects: length of stay (L), acuity of the admission (A), Comorbidity of the patient (measured with the Charlson Comorbidity Index) (C) and emergency department visit in the six months before admission to hospital (E). $C-$statistic with 95\% confidence intervals is used to evaluate the accuracy of the index. The results indicate that LACE Index has moderate discrimination for early death or readmission, therefore it can be used as a measurement to predict risk of patient death and unexpected readmission.

The feasibility and strength of LACE index in predicting 30-day hospital readmission is further studied using anonymised patient data obtained from hospital information system\cite{sd2017paper24}. A positive correlation is observed by univariate logistic regression meaning the increment of LACE index score could greatly result in larger possibility of readmission than elective or day case readmissions (12.8\%, 5.7\% and 1.8\% respectively). In addition, the significance of LACE index is also proved by multivariate logistic regression with an AUC of 0.773 (95\% CI 0.768 to 0.779) and $R^{2}$ of 0.180.

\begin{table}[h]
    \centering
    \small
    \caption{LACE index}
    \scriptsize
    \begin{tabular}{c|c|c|c|c|c}
    \hline
    \hline
    Attribute&Value&Points&Attribute&Value&Points\\
    \hline
    \multirow{7}{*}{L}&\textless1&0&\multirow{4}{*}{C}&1&1\\&1&1&&2&2\\&2&2&&3&3\\&3&3&&$\geq4$&5\\\cline{4-6}&4-6&4&\multirow{5}{*}{E}&0&0\\&7-13&5&&1&1\\&$\geq14$&7&&2&2\\\cline{1-3}
    A&Yes&3&&3&3\\
    \cline{1-3}
    C&0&0&&$\geq4$&7\\
    \hline
    \hline
    \end{tabular}
    \label{tab:table 5}
\end{table}

\textbullet \underline{HOSPITAL Score}, as shown in Table \ref{tab:table 6}, consists of seven independent variables, including hemoglobin at discharge, discharge from an oncology service, sodium level at discharge, procedure during the index admission, index type of admission, number of admissions during the last 12 months, and length of stay. It is derived and validated as a model in a dataset with a total of 12,383 patients discharged from the medical services of the Brigham and Women's Hospital\cite{jd2013hospitalscore}. The seven factors are determined by a series of models: A multi-variable regression model, followed by a regression coefficient-based scoring method and finally a backward multi-variable logistic regression analysis. The internal validation confirms that HOSPITAL score has potential to easily identify patients with a risk of readmission.

An international study validates the HOSPITAL score based on 117,065 adult patients from 9 hospitals across 4 countries\cite{jd2016paper40}. The score is verified in the logistic regression from three aspects: overall performance, discriminatory power and calibration. Overall, the study confirms the discriminatory power of the HOSPITAL score in predicting avoidable readmission, with a C-statistic of 0.72 (95\% CI, 0.72-0.72) and its prediction of potentially readmission matches the observed proportion with an excellent calibration (Pearson $\chi^{2}$ test P=0.89).

\begin{table}[h]
    \centering
    \small
    \caption{HOSPITAL score}
    \scriptsize
    \begin{tabular}{p{6cm}|c}
    \hline
    \hline
    Attributes&Score\\
    \hline
    Hemoglobin $<$ 12g/ml or 120g/L at discharge&1\\
    Discharge from an oncology service&2\\
    Sodium $<$ 135 mEql/L at discharge&1\\
    Procedure performed during hospital stay (any ICD coded procedure)&1\\
    Index admission type: nonelective&1\\
    \# of hospital admission during the previous year&\\
    0&0\\
    1-5&2\\
    \textgreater5&5\\
    Length of stay $\geq$5 days&2\\
    \hline
    \hline
    \end{tabular}
    \label{tab:table 6}
\end{table}

\textbullet \underline{Elixhauser Comorbidity Index}
Because hospital readmission is often tied to complication of disease comorbidity, Elixhauser Comorbidity Index\cite{Elixhauser1998ComorbidityMF} was proposed in 1998 to measure comorbidities in large administrative inpatient datasets. This comorbidity algorithm is developed and tested on the data consisting of 1,779,167 adult patients from 438 acute care hospitals in California. Although original Elixhauser Comorbidity index addresses some limitations from previous measures,  it requires 30 binary variables, which limits its application. Later on, it is revised as a single score for administrative data using backward stepwise multivariate logistic regression\cite{cw2009modify}.

By using Elixhauser Comorbidity index, combined with age and gender date, a hierarchical logistic regression models\cite{wl2020code} predicts 30-day readmission for patients with acute myocardial infarction (AMI), congestive health failure (HF) and pneumonia (PNA). The comparisons using four models, including  1) the hierarchical logistic regression model; 2) XGBoost model with binary ICD-9 codes; 3) a feed-forward ANN model trained on dummy variable representation of ICD-9 code; and 4) an ANN models trained based on latent ICD-9 codes variables, show that the AUC for hierarchical logistic regression is 0.68 (95\% CI 0.678, 0.683) and it is improved by the fourth model to 0.72 (95\% CI 0.718, 0.722).

\textbullet \underline{Baltimore Score (B Score)}
Readmission risk-assessment tools including LACE index, HOSPITAL score and Elixhauser Comorbidity index have been proven effective in predicting patient readmission risk. However, they mainly make predictions based on a portion of patients' features such as length of stay in hospitals and comorbidities. As a result, they may not be able to consider a large number of factors, and some important characteristics may be ignored.

Baltimore score (B score) is a learning based model using thousands of health data variables\cite{dm2019Bscore}. The Baltimore score (B score), invented by researchers from the University of Maryland Medical System (UMMS), is an easily implemented machine learning score to calculate standard readmission risk-assessment scores in real time to predict 30-day unplanned readmissions. This model is individualized for each of three University of Maryland Medical System hospitals in different settings and 382 variables are drew for the final model including demographics, lab test results, \textit{etc.}, from more than 8,000 possible data variables. The research compares the B-score readmission risk level with the actual readmission rates of the three hospitals and the predictions derived from other plans. In the three hospitals, despite the different settings, the overall B score is better than other scores in identifying patients at risk of re-admission, and is the most accurate among the highest-risk patients. The 10\% of patients with the highest risk of B score at discharge have a 30.7\% chance of unplanned readmission. Similarly, the 5\% of patients with the highest B scores at discharge have a 43.1\% chance of being misdiagnosed again. The AUROC of the B score is 0.72 (95\% CI, 0.70-0.73) and it increases to 0.78 (95\% CI, 0.77-0.79) at discharge (all P $<$ .001) compared with the 0.63 (95\% CI, 0.59-0.69) AUROC for HOSPITAL, 0.64 (95\% CI, 0.61-0.68) for Maxim/RightCare, and 0.66 (95\% CI, 0.62-0.69) for modified LACE score. As a result, the study concludes that clinicians can use patient data to calculate B Score and further reduce adverse events and improve patient safety.

\subsubsection{Embedding Features}
Different from singular or hybrid features which measure one or multiple characteristics of objects as one score for evaluation, embedding features (also called latent features) use a feature vector to represent each object. This provides a much more comprehensive capability in describing different types of objects, such as patient, hospitals, diseases \textit{etc.}, than singular or hybrid features.


\begin{figure}[h]
\begin{small}
    \centering
    \includegraphics[scale=0.125]{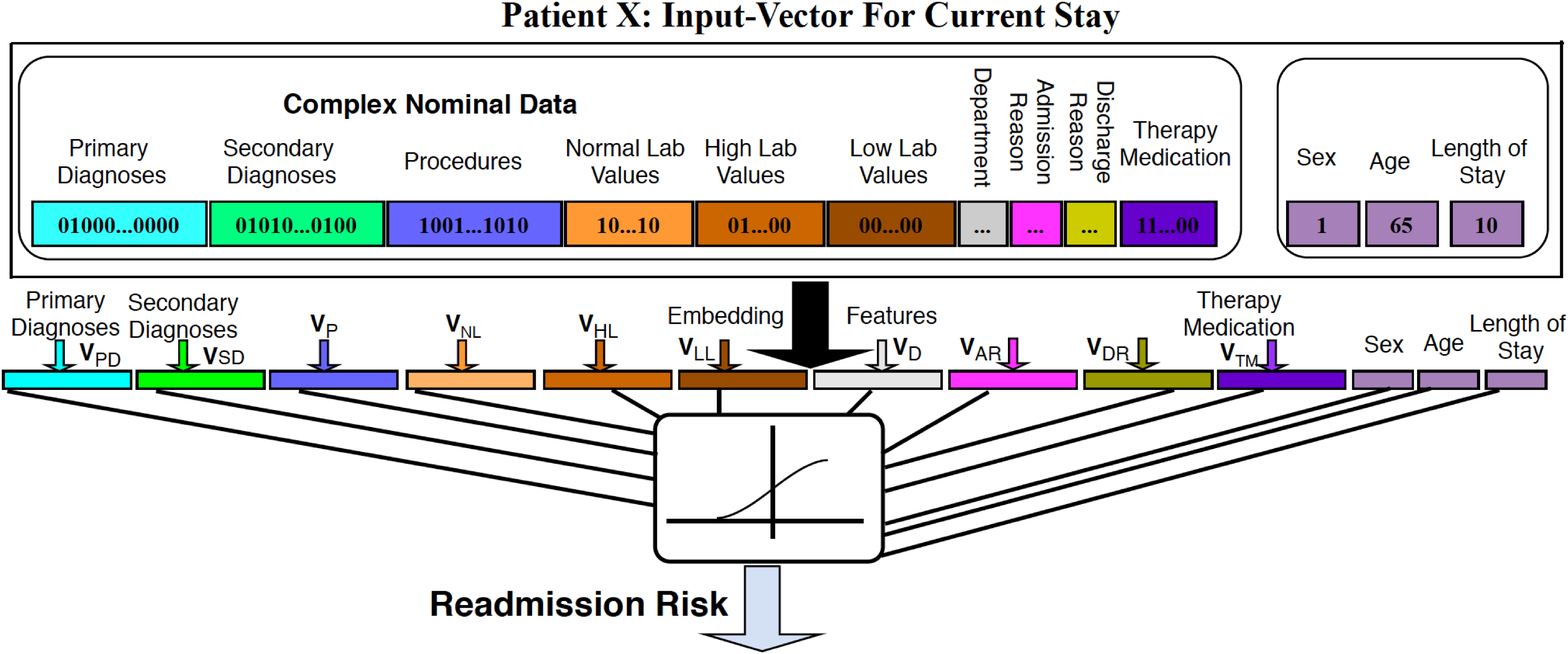}
    \caption{Clinical object embedding based readmission risk prediction framework\cite{dk2014paper26}. Each categorical clinical object, such as primary diagnose, is embedded as a feature vector. Numerical features, such as, age, remain original form. All features are fed into a regression framework to estimate readmission risk.}
    \label{figure:embedding}
\end{small}
\end{figure}

\textbullet \underline{Clinical Object Embedding}
Clinical objects have a variety of types, such as diseases, medical procedures, abbreviations, jargon from doctors, hospitals and other care providers. 
After being admitted to hospital, medical information like diagnoses and procedure codes are recorded as individual features, resulting in a large feature space ineffective for machine learning algorithms to train a model. Alternatively, embedding features can downsize the feature space and simplify the dependencies between input features and target variables.

Instead of using clinical data as features, a latent embedding based framework, as shown in Fig.~\ref{figure:embedding}, is proposed to embed nominal clinical data for hospital readmission prediction\cite{dk2014paper26}. The framework uses latent embeddings of the nominal parts of data, such as diagnosis and procedure codes (ICD-10). An $n\times m$ matrix is organized using clinical data where each of the $n$ rows contains data of a patient during his/her stay in the hospital and each of the $m$ columns denotes one feature. For numerical features, such as age and length-of-stay, the feature are numerical values. For categorical feature, such as diagnose code, one-hot encoding is used to indicate whether a specific code appears in the current visit (1), or not (0). 

To calculate latent embeddings, a data covariance matrix, in Eq.~(\ref{eq:nodesembed}), is first calculated to find relationship between features. In this equation, $X\in\mathbb{R}^{n\times m}$ is an $n\times m$ sparse binary matrix standing for nominal data from the data matrix, $\mu_X$ is the means vector and $\sigma_X$ is the deviations, and $\otimes$ denotes the outer vector product.
\begin{equation}
X_{Cov} = \frac{X^TX-m\cdot(\mu_X \otimes \mu_X)}{(m-1)\cdot\sigma_X\otimes \sigma_X}
\label{eq:nodesembed}
\end{equation}

Given covariance matrix $X_{Cov}$, latent embedding is calculated by applying Singular Value Decomposition to decompose $X_{Cov}$ in Eq.~(\ref{eq:svd}), where $V\in\mathbb{R}^{n\times k}$ is a dense matrix with $n$ rows and $k$ column, meaning each row of the $V$ matrix represents a $k$ dimensional dense feature vector. 
\begin{equation}
X_{Cov} = V\Sigma^2V^T
\label{eq:svd}
\end{equation}

In order to tackle the large number of diagnose code in medical domains, a ``max-pooling'' procedure is applied to select the most responsive latent features from the latent embeddings, so each type of diagnose, such as ``Primary Diagnose'', is denoted by a $k$-dimensional vector. The performance comparison between latent logistic regression and binary logistic regression indicates that latent embedding improves the AUC score of readmission prediction from 0.779 to 0.790.

\textbullet \underline{Patient Embedding}
In addition to embedding each clinical object, research also proposes to represent a patient's entire raw electronic health record (EHR) as a single vector\cite{ar2018patientembed}, or embed a single visit as a feature vector\cite{ec2016medical}. 
By using data from the University of California San Francisco (UCSF) from 2012 to 2016, and the University of Chicago Medicine (UCM) from 2009 to 2016, a research\cite{ar2018patientembed} develops a single data structure representing the EHR dataset in temporal order for each patient. All available data for each patient, from the beginning of a patient’s record until the point of prediction, form the patient’s personalized input to the model to predict unplanned 30-day readmission. A deep learning model is trained and compared with existing EHR and achieves high accuracy for tasks in predicting 30-day unplanned readmission (AUROC 0.75–0.76).

In order to learn efficient expression of medical concepts, Med2Vec\cite{ec2016medical} is proposed to learn healthcare concept at two levels: code-level and visit-level. When learning from code-level information, Skip-gram algorithm is employed to train a non-negative weight ReLU which guarantees sparse code representation production and provides much more convenience to interpret codes. A multi-layer perception network is trained to exploit sequential information of visits $V_t$, presented by a binary vector $\mathbf{x}_t$, and enables prediction of past and future visits. The binary vector can be converted into visit representation as shown in Eq.~(\ref{eq:embedding}) where $\mathbf{W}$ is the code weight matrix and $\mathbf{b}$ is the bias vector.
\begin{equation}
\mathbf{u}_t = ReLU(\mathbf{W}_c\mathbf{x}_t +\mathbf{b}_c)
\label{eq:embedding}
\end{equation}

A visit vector is generated by concatenating visit demographics $\mathbf{d}_t$ with clinical feature vector in Eq.~(\ref{eq:emfinal}). As a result, a patient is represented by concatenating corresponding visit vectors. The performance of Med2Vec is evaluated using data from Children’s Healthcare of Atlanta (CHOA), and shows significant improvement in prediction accuracy.
\begin{equation}
\mathbf{v}_t = ReLU(\mathbf{W}_c[\mathbf{u}_t,\mathbf{d}_t]+\mathbf{b}_v
\label{eq:emfinal}
\end{equation}

\subsection{Domain Segmentation for Data Complexity}
A hospital readmission visit is tied to many factors, including patients, diseases, care facilities, \textit{etc.} 
As a result, it is often ineffective to rely on a one-for-all predictive model for accurate prediction. High readmission rates of hospitals at certain geographical locations might be because of (1) majority patients sharing similar demographics, (2) hospitals mainly practicing certain diseases, or (3) policy and management issues. To tackle such data complexity, one effective way is to rely on domain segmentation to segment readmission learning into multiple small tasks with better data coherence. The purpose is to use patient cohort with less variance (reduced data complexity) to train reliable models. Because readmission rates share rather distinct patterns with respect to different patient groups and diseases, a common approach is to use disease,  gender, or race and ethnicity for segmentation.

\subsubsection{Disease-Specific Prediction}
Disease segmentation is the most common way to tackle the data complexity for readmission prediction. Intuitively, patients suffering from same/similar diseases are likely sharing similar symptoms, treatments, and post-discharge complications. Therefore, using disease to segment patient visits will minimize the data discrepancy for better prediction. 
Many models are proposed to predict readmission for leading cause of death diseases such as hear failure~\cite{vb2015}, stroke~\cite{bk2005race3}, COPD~\cite{cb2017cost}, AMI and pneumonia~\cite{wl2020code}, \textit{etc.}

In a research\cite{vb2015} to determine the risk of unplanned cardiovascular readmission for patients with chronic heart failure, a modified Cox’s proportional hazards model (taking into account the competing risk of death) is used to develop a multivariate prediction model followed by bootstrap methods to identify data factors for the final model. Variables are chosen by a backward-deletion method with a $p$-value threshold for retention. Based on patients information selected from WHICH? trials (Which Heart Failure Intervention is most Cost-effective \& consumer friendly in reducing Hospital care), the $C$-statistic of the final trained model reaches 0.80.

\subsubsection{Gender-Specific Prediction}
Despite the effectiveness of hospital readmission rate control policies and interventions targeted at the whole nation, a gender difference with regard to the rate of 30-day hospital readmissions has been pointed out and this observation triggered a knowledge exploration about gender different in readmission \cite{to2011gender4}. Compared with men, women are reported to have a higher readmission risk after being discharged \cite{ll2018gender1}, especially young women aged under 65 years old \cite{rd2015gender2}. This indicates that policies and interventions aimed at preventing readmissions may need to consider both biological sex and gender in their designs and implementations. A psychiatric administrative dataset with women (n=33,353) and men (n=32,436) patients are used to identify predictors of 30-day readmission. Multi-variable logistic regression models with demographics, clinical information etc. are conducted and the results show female patients have 0.2\% higher risk than male patients to be readmitted within 30 days and suggest that ``Certain key predictors of psychiatric readmission differ by sex''\cite{lb2018gender3}.

\subsubsection{Race and Ethnicity-Specific Prediction}
Race and ethnicity are known factors of health disparity. When it comes to the implementation of HRRP in ethnics, a similar drop in the 30-day readmission rate of Myocardial Infarction patients in both Black and non-Black patients has been previously observed. 
Meanwhile, although the race-specific 30-day readmission rate differs, research also concerns that it might be attributed to patient-level factors such as incoming, age. In other words, the implementation of HRRP has nothing to do with the improvement or deterioration of racial differences in readmission rates and mortality \cite{ap2020race1}. In a mixed-effects logistic regression cohort study of 272,758 adults with diabetes, 
black patients are at a much higher risk of unscheduled readmission to the hospital within 30 days than other racial/ethnic groups with a 12.2\% among black patients, 10.2\% for white individuals, 10.9\% among Hispanic patients and 9.9\% for Asians \cite{rr2019race2}. The higher hospital readmission risk is also confirmed by a research based on 4,784 Blacks and 33,684 Whites with stroke admissions in year 2000 using a truncated negative binomial (TNB) model. The readmission risk among black patients age 65-74 is 40\% higher than white patients \cite{bk2005race3}.

\subsubsection{All-Cause Prediction}
In contrast to the above domain specific approaches, many predictive models still address the hospital readmission using an all-cause approach, which considers all disease types, gender groups, races \textit{etc}. For example, a study compare different models used for readmission risk prediction of hospitalized primary care patients\cite{gg2016comparing}. The study includes four classifiers including Walraven's LACE index, LACE+ index, Donze’s HOSPITAL score and logistic regression based classifiers, to predict 30-day readmission risk, using a dataset at Mayo Clinic Department of Family Medicine. 
The results show that logistic regression based classifiers yield only moderate performance in predicting readmission with a $C$-statistic of 0.666 compared with LACE (0.680), LACE+ (0.662) and HOSPITAL (0.675)\cite{gg2016comparing}.

\section{Model Interpretability}
Model interpretability defines the degree to which the behavior of a model can be explained and understood by human perception. For medical applications, an interpretable and transparent model is almost always preferred. 
Only interpretable predictions allow care providers to understand the decision patterns to support clinical actions. Therefore, the first model level challenge is to design clinically interpretable and practically usefulness models to provide actionable insights to the decision makers. 

\begin{figure}[h]
\begin{small}
    \centering
    \includegraphics[width=85mm,height=50mm]{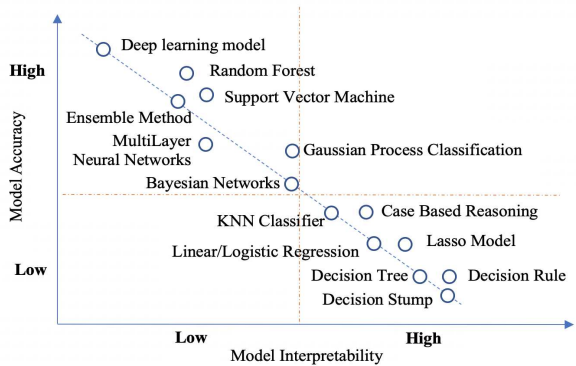}
    \caption{An inverse relationship between model interpretability \textit{vs.} model accuracy for common machine learning algorithms. A model is the output of the referred classifiers (or learning algorithms). In general, a model with a higher accuracy tends to have a lower interpretability, whereas models easier to interpreter tend to be less accurate.}
    \label{figure 3}
\end{small}
\end{figure}

In Fig. \ref{figure 3}, we outline general relationships between model interpretability \textit{vs.} model accuracy of commonly-used machine learning algorithms. The $x$-axis denotes the model interpretability and the $y$-axis denotes the prediction accuracy. In summary, models like decision trees, linear/logistic regressions, case based reasoning \textit{etc.} are relatively easy to interpret, mainly because that the model itself outputs parameters associated to some features or samples. For example, a decision tree will specify features (and their values) used to derive decision, and logistic regress will specify the weight of features for prediction. Case based reasoning finds samples (or prototypes) similar to the query instance to derive decision. As models are moving towards relying on general computing machines (such as support vector machines, neural networks, Bayesian networks) or ensemble models, the interpretability will decrease, mainly because that the models do not directly specify how features/samples are used to form decisions.


\subsection{Model Interpretability Enhancement}

\subsubsection{Simple Transparent Models}
Simple transparent models are the ones with strong interpretability, and the models can directly inform decision actions. Decision trees/rules and linear/logistic regressions, in Fig.~\ref{figure 3}, are the ones falling into this category and frequently being used in the medical domains.  

In a recent study, researchers from MIT\cite{db2020flow} propose to build interpretable predictive models for inpatient flow prediction, which predicts flow of hospitalized patients into two major categories: flows out of the hospital, \textit{i.e.},
discharges, and flows between units of the hospital. A collection of machine learning techniques, including linear models and decision trees, are applied to address four length of stay-related tasks: same-day and next-day discharges, and more-than-7 and more-than-14-day hospital stays. Their study shows that the prediction accuracy and scalability cannot be hindered by attentions on modeling and interpretability. Instead, it brings more interpretable functions and thus clinicians and care providers are more involved, less data as well as computing resources is required.


\subsubsection{Hybrid Transparent Models}
Machine learning algorithms with high accuracy, such as neural networks and support vector machines, often have very little interpretability\cite{mjpaper14}, imposing significant obstacles for clinical decisions and model implementation across health systems\cite{em2020use}. There is a trade-off between performance and interpretability: complexity models are untraceable black boxes, while classic interpretable models are usually simplified with lower accuracy. This trade-off narrows the application of the latest machine learning models in hospital readmission prediction area, which requires both high predictive performance and an understanding of the contribution of each attribute to the model results. 

In order to deliver an interpretable system with higher accuracy, hybrid transparent models combine different types of predictive models to ensure high interpretability and accuracy. A mixture of two types of classifiers, a Boosted C5.0 tree and a support vector machines (SVM), is proposed \cite{lt2016paper3} for hospital readmission prediction. The main objective is to take the advantage of high sensitivity of SVM prediction, as well as the transparency of C5.0 tree model. In their algorithm, two separated routines (optimizing SVM and C5.0 independently, and optimizing mixed ensemble) are carried out in order to determine optimal parameters for SVM and C5.0. A unique feature of the mixed ensemble is that the model delivers a hospital readmission prediction with balanced accuracy and reasonable transparency. 

\subsection{Feature Interpretability Enhancement}
To enhance model interpretability, another effective way is to select/create effective features. By using a small set of important features, the model can inform factors playing important roles for readmission prediction. 

\subsubsection{Feature Selection}
Feature selection selects or extracts significant variables from a set of given features to explain model decisions. 
Such approaches can ensure that the variables included in the final model have clinical significance, can be identified and understood, and can lead to new insights and hypotheses. Most importantly, interpretable machine learning supports clinicians and patients' decision-making by clearly indicating the nature and characteristics of the most important variables for algorithms to make a prediction. 

SHAP (Shapley Additive exPlanations) value\cite{ch2020fs1} is a feature selection based approach to predict personalized patients during and after hospitalization. The interactions between variables are examined by SHAP, followed by Gradient Boosting Machine (GBM) being used to train predictive models based on selected features. The most significant features include primary diagnosis, length of stay, and so on. The final predictive performance is: 30-day readmission (AUC 0.76$/$BSL 0.11); LOS $\geq$ 5 days (AUC 0.84$/$BSL 0.15); death within 48–72h (AUC 0.91$/$BSL 0.001). Similarly, information-based (Gini indexing) and frequency-based feature reduction methods are used to improve the understandability of determining readmissions based on a number of variables. Their research shows that prescription drugs, diagnostics, and information about operations during admission can help predict the admission rate\cite{a2013}.

\subsubsection{Feature Learning}
Feature learning intends to transform complex, redundant, and high dimensional features into a new representation, such as a dense feature vector, that can be effectively exploited in machine learning tasks. The strength of feature learning is that it extracts useful features or representations from raw data, and also learns the predictive task at the same time. Feature learning can be divided into two categories: supervised feature learning in which features are learned with labeled input data and unsupervised feature learning, which learns features from unlabeled input data.

For most feature learning methods, the output features are dense vectors suitable for machine learning, but difficult for human understanding. A recent study\cite{zh2019vectors} proposes to embed 971 features into 100 sparse dimensions through a $k$-sparse auto-encoding using health data records from Yale New Haven Health system. $k$-sparse auto-encoder is a sparse embedding function which carries the most vital information for each patient and considers the embedded vectors as a concentration of the original features. Given a dataset $\mathcal{D}=[\mathsf{X}_1,\mathsf{X}_2,\cdots,\mathsf{X}_n]\in\mathbb{R}^{n\times m}$ with $n$ instances and $m$ original features, and each instance $\mathsf{X}_i \in\mathbb{R}^{m}$ is denoted by an $m$ dimensional features, the classical autoencoder objective function is defined as Eq. (\ref{eq: auto}). $\phi$ denotes encoder transition converting $\mathsf{X}_i$ to a new $k$ dimensional feature space $\widetilde{\mathsf{X}}_i\in\mathbb{R}^{k}$, and $\psi$ is the decoder transition converting $\widetilde{\mathsf{X}}_i$ back to the original input space. $(\phi\circ\psi)\mathsf{X}_i$ denotes the composite function, encoding and decoding to $\mathsf{X}_i$. 
\begin{equation}
\begin{aligned}
   \phi : \mathsf{X} \rightarrow \widetilde{\mathsf{X}}\qquad\psi : \widetilde{\mathsf{X}} \rightarrow\mathsf{X}\\\argmin\limits_{\phi,\psi}\frac{1}{\left|\mathcal{D}\right|}\sum\nolimits_{i=1}^{n} \|\mathsf{X}_i-(\phi\circ\psi)\mathsf{X}_i\|_{2}^{2}
\end{aligned}
\label{eq: auto}
\end{equation}
Then a $k$-sparse activation constraint in Eq. (\ref{eq: constraint}) is used to penalize the deviation between observed average activation value of neurons $\rho_{h,\mathcal{D}}$ from pre-defined average activation value $\rho_{h,\mathcal{D}}^{*}$ on all instances in $\mathcal{D}$. 
\begin{equation}
\begin{aligned}
\argmin\limits_{\phi,\psi} \sum\nolimits_{h\in\phi,\psi} \|max(0,\rho_{h,\mathcal{D}} - \rho_{h,\mathcal{D}}^{*})\|^{2}
\end{aligned}
\label{eq: constraint}
\end{equation}
Finally, a secondary constraint is applied to limit each dimension of the sparse embedding to either 1 or 0. This binarization forces the sparse embedding to carry the most significant information for each patient. 
\begin{equation}
\begin{aligned}
\argmin\limits_{\phi,\psi}\frac{1}{\left|\mathcal{D}\right|}\sum\nolimits_{i=1}^{n}\sum\nolimits^{H}_{h=1}(\widetilde{\mathsf{X}}_{i,h}\times(1-\widetilde{\mathsf{X}}_{i,h}))
\end{aligned}
\end{equation}

Based on the above design, a hierarchical structure in an interpretable framework is embedded with the sparse vectors.  The patient data distribution shows that sparse embedding can make the hidden phenotype in the admission cohort more prominent. Therefore, the method shows capability of maintaining the integrity of important information.

\subsection{Feature-Model Interpretability Enhancement}
Medical data often have high dimensions, resulting in low learning efficiency, high complexity models with poor interpretability\cite{mjpaper14}. To enhance the interpretability, feature and model combined enhancement approaches integrate feature learning and model training together to learn features easy to interpret and models with a better prediction accuracy.

\subsubsection{Feature Regularization Models}
Sparsity regularization is commonly used to regularize the feature space by using different weight values for the predictive models to leverage feature weights for prediction. 

\textbullet \underline{Sparsity Regularization} Sparsity regularization uses L$_1$ norm normalization to enforce the sparsity, so majority features have small weight values except that a few important features have relatively large values. In the research aiming to identify unplanned readmission risks in a high dimensionality electronic health dataset, a tree-lasso logistic regression \cite{mjpaper14} is proposed to integrate ICD-9-CM hierarchy for pediatric hospital readmission. The learning objective is to minimize a penalized likelihood defined in Eq.~(\ref{eq:tree-lasso}).
\begin{equation}
\mathcal{L}(\theta)=\sum_{i=1}^{n}(1+\exp(-y_i(\mathbf{x}_i^T\theta+c)))+\lambda P(\theta)
\label{eq:tree-lasso}
\end{equation}
In Eq.~(\ref{eq:tree-lasso}), $\theta$ is the model parameter (weight values of features), $\mathbf{x}_i$ defines feature vector of a patient (such as ICD-9-CM codes of each visit), and $y_i\in\{-1,1\}$ is the label of $\mathbf{x}_i$. $c$ and $\lambda$ are predefined parameters. $P(\theta)$ defines the regularization term, and for Lasso regularization it is defined as $P(\theta)=\sum_{j=1}^m|\theta_j|$ which encourages to select ``sparse'' features. In reality, Lasso regularization still selects a rather large number of features hard to interpret, so Tree-Lasso regularization uses another regularization term defined in Eq.~(\ref{eq:tree-lasso-regularization}). The key idea is to utilize the ICD-9-CM hierarchy of diagnoses, which represents groups of diagnostic does. For example, code between 001 and 139 represent infectious and parasitic diseases, and 001 to 009 further represent intestinal infectious diseases. 
\begin{equation}
P(\theta)=\sum_{j,k}w^k_j||\mathbf{x}_{G_k}||_2
\label{eq:tree-lasso-regularization}
\end{equation}
In Eq.~(\ref{eq:tree-lasso}), $G_k$ defines the group of features selected by the $k^{th}$ node in the hierarchy. $||\cdot||_2$ is the L$_2$ norm, and $w^k_j$ is the weight assigned to each node. 

The above model was evaluated using 66,000 pediatric hospital records, in terms of model interpretability and selecting features on different hierarchical levels of disease classes. Comparing to traditional Lasso logistic, the integration of ICD-9-CM hierarchy diseases provides better model interpretability and AUC values (0.779) with less information than traditional Lasso model.

Domain hierarchical structure of the diseases code (\textit{e.g.} ICD-9-CM hierarchy) is also used in another tree based sparsity-inducing regularization\cite{jj2018sr1} which exploits domain induced hierarchical structure for disease codes to improve the comprehension of hospital readmission topic. In their study, four regularizaition methods are compared in respect to logistic regression model. Sparse Group Regularizer (SGL) and Tree Structures Group Regularizer (TSGL) outperform L$_1$ and L$_2$ norm regularization. SGL assumes the input features can be set into $k$ groups while TSGL applies the hierarchical structure presented on the features. Also, the TSGL method is proved with better sparsity at higher levels of the hierarchy. 

\textbullet \underline{Attention Regularization} Attention is a neural network mechanism to assign weight values to features/instances, by taking their correlation into consideration. An attention based neural network model (MLP\_attention) is proposed to generate interpretable prediction results to determine the contribution of each feature. The final input feature representation is calculated by element-wise multiplication of input feature vector $X$ and the attention signal $\alpha$ generated by a fully connected layer\cite{ppc2020sr2}.

A two-level neural attention model\cite{ce2016retain} proposes to consider hospital visits in a reverse time order so that recent clinical visits receive higher attention weights. The model, validated on 14 million  visits, delivers better accuracy and interpretability compared to generic deep learning methods. It can also detect influential past visits and significant clinical variables within those visits (e.g. key diagnoses).

\subsubsection{Feature Topic Models}
Topic modeling is a machine learning approach that explores high level semantics from text documents. In a general setting, a set of documents are used to find clusters of words with similar semantics, which are interpreted as topics. Intuitively, feature vectors representing patients' in-hospital treatment can be very sparse and the relationships between features with similar meanings are also ignored. Topic modelling is one effective way to solve this challenge, by assigning features to different topics. This will essentially enhance the interpretability for both model and features, because users can understand which features, such as ICD codes, are tied to certain disease (topics).

\begin{figure}[h]
\begin{small}
    \centering
    \includegraphics[width=3.0in,height=1.0in]{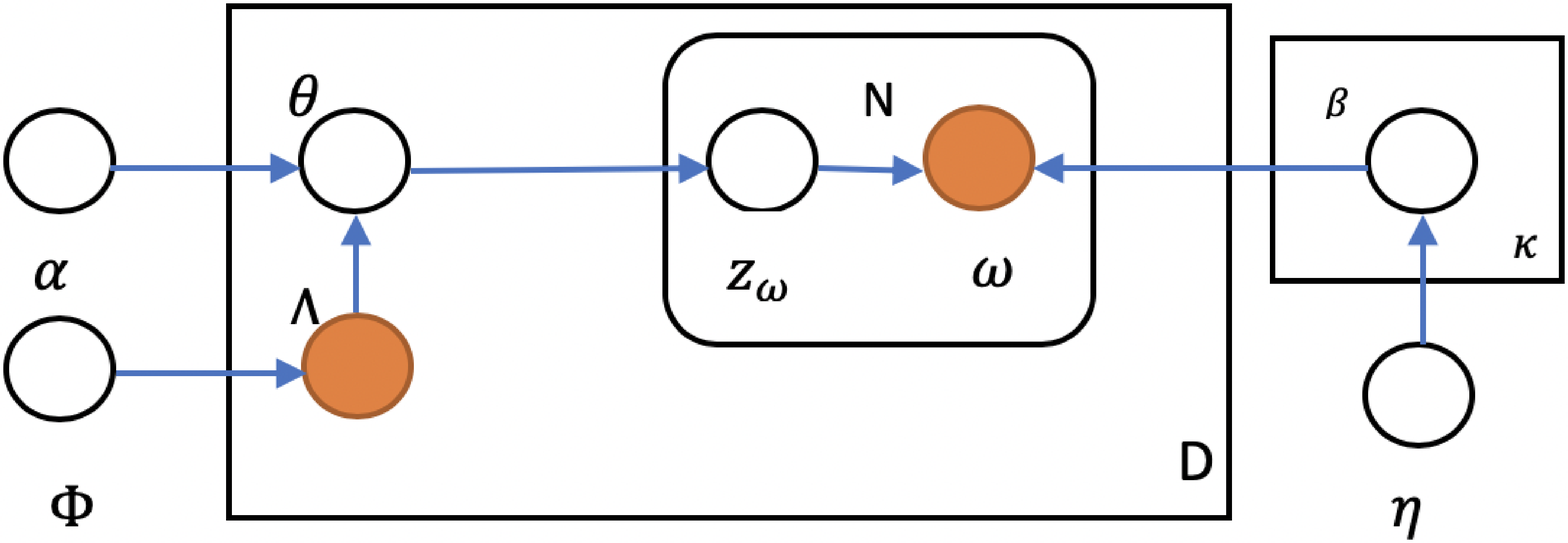}
    \caption{Labeled-LDA topic model. Colored nodes are observable variables, where $\omega$ is a list of word indices (there are $N$ words in total), and $\Lambda$ is a list of binary topic presence/absence indicators. $\alpha$ is the parameter of the Dirichlet topic prior; $\phi$ is the labeling prior probability; $\theta$ is the mixture distribution over all topics; $\mathnormal{z}_{\omega}$ is the word-topic assignments; $\beta$ is the multi-nomial topic distributions over vocabulary; $\eta$ is the parameters of the word prior.}
    \label{figure:lda}
\end{small}    
\end{figure}

Latent Semantic Analysis (LSA), Hierarchical Dirichlet Process (HDP), and Latent Dirichlet Allocation (LDA) are commonly used topic models, but they normally cannot take label information into consideration to learn topics. 
In a research to predict post-ICU mortality\cite{yl2017tf1}, labeled-LDA (Latent Dirichlet Allocation)\cite{dr2009lda} is used to incorporate domain knowledge into topic modeling, by using ICD-9-CM codes as labels (domain knowledge). Given a dataset with $D$ documents and $N$ unique words ($\omega_1,\omega_2,\cdots,\omega_N$), a list of binary topic presence/absence indicators $\Lambda^{(d)}=(l_1, . . . , l_K)$ are used to denote topic(s) the current document $d$ is associated to. The learning of labeled-LDA\cite{dr2009lda} is to train a generative model in Fig.~\ref{figure:lda}. Unlike traditional LDA model in which a multinomial mixture distribution $ \theta^{(d)}$ is drawn over a ${K}$ topics for each document $\mathnormal{d}$ from a Dirichlet prior $\alpha=(\alpha_1,\cdots,\alpha_{{K}})$, in Labeled-LDA, the multinomial mixture distribution $ \theta^{(d)}$ is restricted only over topics that equivalent to its labels $\Lambda ^{(d)}$ as shown in Fig.~\ref{figure:lda}. 

Table \ref{table 7} lists examples of ICD-9-CM codes, their definitions along with the corresponding keywords learned from Labeled-LDA. For example, ``Anemia'' topic includes words ``tablet'', ``mg'', ``blood'', ``po'', \textit{etc.} There are two advantages of using the ICD-9-CM code as a label in Labeled-LDA. First, the clinical notes from a given patient's record only contribute to a subset of topics corresponding to that patient's ICD-9-CM code assignment. Secondly, by combining the ICD-9-CM code definition and the keywords of a given theme, the interpretability of the theme can be realized.

\begin{table}[h]
    \centering
    \small
    \caption{Examples of feature topic models using Labeled-LDA to find keywords associated to ICD-9-CM codes. ``Definition'' shows diseases, injuries, symptoms and conditions defined by ICD-9-CM codes. ``Keywords'' are the words learned to correspond to the ICD-9-CM.}
    \scriptsize
    \begin{tabular}{p{1.2cm}|p{1.8cm}|p{5cm}}
    \hline
    \hline
    ICD-9-CM&Definition&Keywords\\
    \hline
    280-285&Anemia&pt tablet mg blood po hct sig daily discharge pm doctor namepattern md patient pain day history gi admission hospital\\
    \hline
    420-429&Other forms of heart disease&pt mg patient hr chest resp left lasix gi po stable pain gu neuro gtt bp bs day cv plan\\
    \hline
    270-279&Other metabolic and immunity disorders&patient mg pt chest day left artery pain po stable coronary cabg status discharge history date post namepattern clip examination\\
    \hline
    317-319&Mental retardation&pt tube noted chest cc resp patient retardation thick secretions care cont plan trach abd hr ct neuro telemetry coarse\\
    \hline
    \hline
    \end{tabular}
    \label{table 7}
\end{table}

By using textual information, such as medical notes and other structured data from MIMIC \uppercase\expandafter{\romannumeral2} database, 
the model is validated to achieve transparent outcome prediction and high AUC scores (0.835 and 0.829) for 30-days and 6-months readmission prediction, respectively\cite{yl2017tf1}.

\subsubsection{Feature Interaction Models}
Using interactions between variables can interpret models by explaining factors playing important roles in the decision. Accordingly, generalized additive models with pairwise interactions (GA$^{2}$Ms) is proposed using standard Generalized additive models (GAMs), which usually model the dependent variable as a sum of univariate models \cite{ly2013}. Pairwise interactions is added into GAMs in order to improve model accuracy and interpretability. Eq.(\ref{eq:gams}) shows how GA$^{2}$Ms works: $x_i = (x_{i1},...,x_{ip})$ is the feature vector with $p$ features and $y_i$ is the target. $g$ is the link function, for each item $f_j$, $E[f_j] = 0$. For GA$^{2}$Ms, it builds the best GAM at first and then all possible pairs of interactions in the residuals are detected and ranked. It is tested on a readmission prediction for pneumonia case study and GA$^{2}$M models present an excellent readmission prediction accuracy while at the meantime, the interpretability of the decisions is improved. GA$^{2}$Ms is proved to be useful in accurately explaining the prediction decisions made by model for each patient focusing on the most relevant terms for each patient \cite{cr2015gams}.  
\begin{equation}
    g(E[y]) = \beta_0 + \sum_{j}f_{j}(x_j)+ \sum_{i\neq j}f_{ij}(x_i,x_j)
    \label{eq:gams}
\end{equation}

Using deep neural networks, INPREM\cite{zx2020inprem} is recently proposed to enhance deep learning interpretability for interpretable and trustworthy healthcare prediction. To ensure model interpretability, INPREM employs a $CM\in\mathbb{R}^{T\times|C|\times l}$ matrix to characterize contributions of different factors, where $T$ denotes the number of hospital visits, $|C|$ denotes number of medical/procedure codes, and $l$ denotes the number of classes. $CM[i,j][k]$ captures the contribution of the $j^{th}$ medical event in the $i^{th}$ visit, with respect to the class $k$. By using deep neural networks to learn weight matrices $\mathbf{w}_v\in\mathbb{R}^{g\times|C|}$ and $\mathbf{w}_c\in\mathbb{R}^{g\times l}$ where $g$ is the dimension of the embedding space. It also uses $\alpha\in\mathbb{R}^{1\times T}$ to capture non-linear dependencies between visits, and uses $\beta\in\mathbb{R}^{g\times T}$ to encode non-linear dependencies between medical events within each visit, INPREM calculates CM matrix using Eq.~(\ref{eq:inprem}), where $\odot$ denotes element-wise multiplication. 
\begin{equation}
CM[i,j]=\mathbf{W}^{T}_c(\alpha[i]\beta[:,i] \odot \mathbf{W}_v[:,j])
    \label{eq:inprem}
\end{equation}


\section{Model Implementation Conflict}
HRRP is evidently effective in encouraging hospitals, especially the ones with high readmission rates, to reduce readmission rates\cite{jw2017}. Ideally, it is expected that the program will motivate hospitals to improve treatments, enforce discharge evaluation and follow-up procedures, and eventually lead to medical cost reduction and improve the patient satisfaction and qualify of services\cite{rz2016observation}. In reality, however, the implementation of the models is challenged by many conflicts, partially because predictive models create a layer of separation between patients and care providers.

\subsection{Quality Contradiction}
\subsubsection{Controversial Measurement}
Concerns about the HRRP program, as a healthcare service quality measurement, have been rising due to the complexity of the healthcare ecosystem. One one hand, for many patients, even if they do receive high-quality care, there still exists possibility the patients will return to the hospital due to the nature of disease development. An inverse relationship between a hospital’s readmission rate and its mortality rate has been observed, and shows that low-mortality hospitals tend to have higher readmission rates\cite{aj2018tofix}. On the other hand, prediction models usually fail to include data elements that have perverse incentives such as prior utilization. In order to determine whether the healthcare expenses of patients with a 30-day readmission differ from those without a 30-day readmission, a retrospective study is designed to measure the consequences. After comparing the 12 month genuine costs of two groups of patients: patients with a 30-day readmission \textit{vs.} patients without 30-day readmission, the study show that readmitted patients are often ``sicker'' than non-admitted patients. Therefore, using hospital readmission model for profiling hospitals may  ``systematically underestimate performance of hospitals with high rates of observed readmissions'', and jeopardy the true ``saving'' to the healthcare system\cite{asz2019cm1}.

Another research\cite{apc2014cost} also concludes that certain patient groups, such as cardiac arrest, have high readmission rates and inpatient costs. A two-part model, including a logistic model and a gamma regression model,  is constructed to determine adjusted costs and cost ratios based on a cohort from GWTG-Resuscitation submitted by 523 acute-care hospitals with a total number of 19,373 patients. The overall amount of 30-day readmission is 2005 in which cardiovascular disease contributes to 35.9\% followed by pulmonary disease (17.1\%). For inpatient resource usages, the mean cost of in-hospital cardiac arrest is \$35,808 $\pm$ \$38,230, much higher than the average cost \$7,741 $\pm $ \$2,323 of the whole cohort.

\subsubsection{Unintended Consequence}
After the HRRP initiative, the readmission rate has shown an annually downward trend. Yet, the true factors and the consequence of this reduction have not been clearly understood so far. One possible reason is that hospitals choose to avoid readmissions by raising the threshold to readjust patients recently discharged from the hospital. A recent research\cite{ch2020assessment} studies the HRRP policy on the emergency (ER) visit, and observes that HRRP is associated with a decreasing likelihood ``that hospitals would readmit recently discharged patients returning to the ED within 30 days of discharge''. Such observations raise concerns about unintended consequence of the HRRP program, and its potential long term impact to both hospitals and patients. 

\subsection{Cost Contradiction}
\subsubsection{Emotional Cost Implication}
Being readmitted into hospital after a recent discharge is a complicated experience for many patients. Researchers find that patients often harbor different opinions towards their own readmission from healthcare providers\cite{am2019ec1}. The research, from 178 interviews of readmitted patients compared with the perspective of the admission provider, concludes that 58\% patients believe that system issues,  controlled by the discharge process, are the contributors to their readmission. Meanwhile, patients concern that their readmission could have been avoided if the system were modified. In NSW Australia, in order to explore the readmission experience of aged population, three elderly patients discharged from a large tertiary referral hospital are interviewed. Their feedback suggests that nurses are more proactive in people-centered care of older patients to alleviate the patient being left out and let down\cite{sd2012ec2}. 

In order to study relationship between patient experience and predictive modeling of readmission, a research\cite{zs2018ec3} uses patient-level Hospital Consumer Assessment of Healthcare Providers and Systems (HCHAPS) and Press Ganey data to understand whether a patient being readmitted is a driver of poor experience scores (reverse causation). The study uses multivariable logistic regression to analyze how HCHAPS and Press Ganey data 30-readmission status are correlated. The results point out that ``poor patient experience may be due to being readmitted, rather than being predictive of readmission''. In other words, predictive models do not directly lead to poor patient experience\cite{zs2018ec3}. 

\subsubsection{Financial Cost Implication}
Despite of the cost reduction motivation of the HRRP initiative, the total costs to hospitals and patients are contracting. At very minimum level, the implementation of predictive models in hospitals, prior to the discharge of patients for preventable readmission, is time consuming and expensive. The long term accumulated costs can even be more profound. A study using hierarchical regression analyses to examine the cumulative costs of short-term readmissions after percutaneous coronary intervention (PCI) shows that 30 days of readmission and cumulative expenses are positively correlated\cite{at2017thirty}. To reduce readmission, a study suggests hospitals to consider providing better medication reconciliation, giving instruction to both patients and their caregivers on the follow-up care and also implementing post-discharge care coordination\cite{em2020use}. The post-discharge education and follow-up intervention include complimentary home visit within 48 hours after a patient is discharged to review the education provided before discharge and a minimum of four nursing telephone meetings. Despite of the fact that the intervention\cite{mh2013tohome} on a patient after discharge is effective in reducing readmission rate, it imposes significant extra costs to medical staff and healthcare systems.

A research \cite{rw2018associa} investigates 
8 million Medicare beneficiary hospitalizations from 2005 to 2015, and finds that the implementation of the HRRP is associated with a significant increase in 30-day postdischarge mortality among beneficiaries hospitalized for heart failure and pneumonia. In other words, there is an association that 30-day readmission policy may increase the mortality for patients beyond the 30 day window. On the other hand, another study\cite{mb2020cost} shows that transitional care interventions, such as home visits by nurses, can reduce death rates and hospital readmissions by more than 30\%. As a result, they suggest that transitional care services should become the standard of care for post-discharge management of patients with heart failure.

\subsection{Intervention Contradiction}
\subsubsection{Patient Willingness}
The implementation of predictive models implies an intervention process which may contradict to patients' willingness. For example, a model may predict a patient as high readmission risk and requires further hospitalization, but the patient feels fine and is unwilling to cope. Discharge against medical advice (AMA) means a patient chooses to leave the hospital earlier than the discharge recommendation. Leaving the hospital without following the doctor’s advice may put the patient at risk of insufficient medical care and result in the need for readmission. A retrospective cohort analysis based on 2014 National Readmission Database shows that patients discharged against medical advice (AMA) are of larger possibility of 30-day readmission compared with non-AMA individuals, which has caused huge loss to the healthcare system and higher hospital mortality. Patients discharged from the AMA are also more likely to have an earlier rebound and readmission, which may reflect their dissatisfaction with the initial care\cite{st2020association}.
\subsubsection{Patient Privacy}
As EMRs are becoming common, the automatic collection, use and storage of patient medical information have been facilitated, allowing studies on hospital readmission and stimulating care providers for automated decision making. However, the applications of large amounts of medical data also raise public concerns about patients' medical information leakage. Although HIPAA and other institutions prohibit researchers to use identified patient-level data and patients’ personal information are protected properly by this policy\cite{45cr2012,2013keguide}, indeed, there are always risks of information breach. Developers need to find a balance between collecting sufficient representative patients’ medical data and minimizing the risk of leaking patients’ privacy. A governance structure that includes patients and other stakeholders early in the development of a model is recommended to be implemented by model developers at the earliest stage of development\cite{ic2014}. 

\section{Datasets and Resources}
In this section, we summarize a list of datasets publicly available for hospital readmission prediction in Table \ref{table8}. 
\subsection{All Cause Readmission Datasets}
All cause datasets 
contain information about all hospitalized patients of typical diseases within a specific period of time. 

Among all data sources, the Healthcare Cost and Utilization Project (HCUP) contributes significantly to the nationwide databases in healthcare readmission. HCUP is developed as a combination of healthcare databases, software tools, and products through federal-state-industry partnerships and are jointly developed by healthcare research and the Agency for Healthcare Research and Quality (AHRQ). Table \ref{table 9} lists the National Inpatient Dataset (NIS), State Inpatient Databases (SID), and Nationwide Readmissions Database (NRD), which all belong to the HCUP databases.  

The HCUP database brings together the data collection efforts of state data organizations, hospital associations, private data organizations, and the federal government to create a national information resource for encounter-level medical data  as shown in Table \ref{table 9}. HCUP includes the largest collection of longitudinal hospital care data in the United States, starting from 1988 with full payment, encounter level information. These databases can be used to study a wide range of health policy issues, including the cost and quality of medical services, medical practices at the national, state and local market levels, access to health care plans, and treatment outcomes.

\begin{table}[h]
    \centering
    \caption{HCUP Databases}
    \scriptsize
    \begin{tabular}{p{0.2\textwidth}|p{0.16\textwidth}|p{0.05\textwidth}}
    \hline
    \hline
    Dataset&Description&\# of records\\
    \hline
    National (Nationwide) Inpatient Sample (NIS)
    \href{https://www.hcup-us.ahrq.gov/db/nation/nis/nisdbdocumentation.jsp}{\nolinkurl{https://www.hcup-us.ahrq.gov/}}
    \href{https://www.hcup-us.ahrq.gov/db/nation/nis/nisdbdocumentation.jsp}{\nolinkurl{db/nation/nis/nisdbdocumen}}
    \href{https://www.hcup-us.ahrq.gov/db/nation/nis/nisdbdocumentation.jsp}{\nolinkurl{tation.jsp}}& U.S. regional and national estimates of inpatient utilization, access, charges, quality, and outcomes&7 million each year\\
    \hline
    Kids' Inpatient Database (KID)
    \href{https://www.hcup-us.ahrq.gov/db/nation/kid/kiddbdocumentation.jsp}{\nolinkurl{https://www.hcup-us.ahrq.gov/}}
    \href{https://www.hcup-us.ahrq.gov/db/nation/kid/kiddbdocumentation.jsp}{\nolinkurl{db/nation/kid/kiddbdocumen}}
    \href{https://www.hcup-us.ahrq.gov/db/nation/kid/kiddbdocumentation.jsp}{\nolinkurl{tation.jsp}}&The largest publicly-available all-payer pediatric inpatient care database in the US&3 million each year\\
    \hline
    Nationwide Ambulatory Surgery Sample (NASS)
    \href{https://www.hcup-us.ahrq.gov/db/nation/nass/nassdbdocumentation.jsp}{\nolinkurl{https://www.hcup-us.ahrq.gov/}}
    \href{https://www.hcup-us.ahrq.gov/db/nation/nass/nassdbdocumentation.jsp}{\nolinkurl{db/nation/nass/nassdbdocumen}}
    \href{https://www.hcup-us.ahrq.gov/db/nation/nass/nassdbdocumentation.jsp}{\nolinkurl{tation.jsp}}&The largest all-payer ambulatory surgery database in the United States&9.9 million each year\\
    \hline
    Nationwide Emergency Department Sample (NEDS)
    \href{https://www.hcup-us.ahrq.gov/db/nation/neds/nedsdbdocumentation.jsp}{\nolinkurl{https://www.hcup-us.ahrq.gov/}}
    \href{https://www.hcup-us.ahrq.gov/db/nation/neds/nedsdbdocumentation.jsp}{\nolinkurl{db/nation/neds/nedsdbdocumen}}
    \href{https://www.hcup-us.ahrq.gov/db/nation/neds/nedsdbdocumentation.jsp}{\nolinkurl{tation.jsp}}&The largest all-payer emergency department (ED) database in the United States&30 million each year\\
    \hline
    Nationwide Readmissions Database (NRD)
    \href{https://www.hcup-us.ahrq.gov/db/nation/nrd/nrddbdocumentation.jsp}{\nolinkurl{https://www.hcup-us.ahrq.gov/}}
    \href{https://www.hcup-us.ahrq.gov/db/nation/nrd/nrddbdocumentation.jsp}{\nolinkurl{db/nation/nrd/nrddbdocumen}}
    \href{https://www.hcup-us.ahrq.gov/db/nation/nrd/nrddbdocumentation.jsp}{\nolinkurl{tation.jsp}}&Support various types of analyses of national readmission rates for all patients
    &18 million each year\\
    \hline
    State Inpatient Databases (SID)
    \href{https://www.hcup-us.ahrq.gov/db/state/siddbdocumentation.jsp}{\nolinkurl{https://www.hcup-us.ahrq.gov/}}
    \href{https://www.hcup-us.ahrq.gov/db/state/siddbdocumentation.jsp}{\nolinkurl{db/state/siddbdocumen}}
    \href{https://www.hcup-us.ahrq.gov/db/state/siddbdocumentation.jsp}{\nolinkurl{tation.jsp}}&Inpatient discharge records from community hospitals in that State&State-specific\\
    \hline
    State Ambulatory Surgery and Services Databases (SASD)
    \href{https://www.hcup-us.ahrq.gov/db/state/sasddbdocumentation.jsp}{\nolinkurl{https://www.hcup-us.ahrq.gov/}}
    \href{https://www.hcup-us.ahrq.gov/db/state/sasddbdocumentation.jsp}{\nolinkurl{db/state/sasddbdocumen}}
    \href{https://www.hcup-us.ahrq.gov/db/state/sasddbdocumentation.jsp}{\nolinkurl{tation.jsp}}&Encounter-level data for ambulatory surgeries and may also include various types of outpatient services&State-specific\\
    \hline
    State Emergency Department Databases (SEDD)
    \href{https://www.hcup-us.ahrq.gov/db/state/sedddbdocumentation.jsp}{\nolinkurl{https://www.hcup-us.ahrq.gov/}}
    \href{https://www.hcup-us.ahrq.gov/db/state/sedddbdocumentation.jsp}{\nolinkurl{db/state/sedddbdocumen}}
    \href{https://www.hcup-us.ahrq.gov/db/state/sedddbdocumentation.jsp}{\nolinkurl{tation.jsp}}&Emergency visits at hospital-affiliated emergency dept. (EDs) not resulting in hospitalization&State-specific\\
    \hline
    \hline
    \end{tabular}
    \label{table 9}
\end{table}

\begin{table*}[h]
    \centering
    \caption{Public datasets used for hospital readmission prediction (Demographics (Demo.), Hospital Information (Hosp.), Clinical Information (CL), Healthcare utilization (HC)}
    \scriptsize
    \begin{tabular}{p{6.3cm}|p{1.6cm}|p{1.5cm}|l|p{1.3cm}|l|p{0.5cm}|p{0.5cm}|p{0.5cm}}
    \hline
    \hline
    Dataset \& Paper &Type&Population&\#of patients&\#of admin.&Demo.&Hosp.&CL.&HC.\\
    \hline
    Medicare Provider Analysis \& Review (MedPAR)\cite{tsai2014,sf2015paper71}&All cause&Medicare&10E5&NA&\Checkmark&\Checkmark&\Checkmark&\Checkmark\\
    \hline
    Medicare Current Beneficiary Survey (MCBS)\cite{asz2019cm1}&All cause&Medicare &10E5&NA&\Checkmark&\Checkmark&\Checkmark&\Checkmark\\
    \hline
    Health Care’s Enterprise Data Warehouse (EDW)\cite{sa2013}&All cause&18+&10E5&NA&\Checkmark&\Checkmark&\Checkmark&\Checkmark\\
    \hline
    Cerner’s Millennium® EHR software system\cite{sa2013}&All cause&NA&10E5&NA&\XSolid&\XSolid&\Checkmark&\XSolid\\
    \hline
    New Zealand National Minimum Dataset\cite{v2012,jf2015comparison}&All cause&All patients&10E5&NA&\XSolid&\XSolid&\Checkmark&\XSolid\\
    \hline
    National Inpatient Dataset (NIS)\cite{kz2013}&All cause&All patients&10E5&NA&\Checkmark&\Checkmark&\Checkmark&\Checkmark\\
    \hline
    State Inpatient DB (SID)\cite{vm2015,kz2015paper53,jh2014paper95,mjpaper14,sr2015paper21,lb2018gender3}&All cause&All patients&10E5&NA&\Checkmark&\XSolid&\Checkmark&\Checkmark\\
    \hline
    Nationwide Readmissions Database (NRD)\cite{wl2020code}&All cause&All patients&10E5&NA&\Checkmark&\Checkmark&\Checkmark&\Checkmark\\
    \hline
    Resource and Patient Management System (RPMS)\cite{ck2018paper36}&All cause&All patients&10E5&NA&\Checkmark&\XSolid&\Checkmark&\Checkmark\\
    \hline
    Nationally Rep. Health and Retirement Study (HRS)\cite{mb2015paper47}&All cause&50+&20000&NA&\Checkmark&\XSolid&\Checkmark&\XSolid\\
    \hline
    Queensland Hosp. Admit. Patient Data (QHAPDC)\cite{sw2009paper96}&All cause&All patients&10E5&NA&\Checkmark&\XSolid&\Checkmark&\XSolid\\
    \hline
    MIMIC III database\cite{arbj2019machine}&All cause&16+&38,597&49,785&\Checkmark&\XSolid&\Checkmark&\Checkmark\\
    \hline
    MIMIC II database\cite{yl2017tf1}&All cause&16+&26,000&NA&\Checkmark&\XSolid&\Checkmark&\Checkmark\\
    \hline
    Cerner Health Facts database\cite{bk2018}&All cause&All patients&17,880,231&74,036,643&\Checkmark&\Checkmark&\Checkmark&\Checkmark\\
    \hline
    Natl. Surgical Quality Imp. Program\cite{rm2015underlying,lg2014paper30,mk2012paper81}&Surgery&All patients&NA&NA&\Checkmark&\Checkmark&\Checkmark&\Checkmark\\
    \hline
   PREVENT III database\cite{jm2013paper85}&Limb ischemia&18+&1404&24.4\%&\Checkmark&\XSolid&\Checkmark&\Checkmark\\
   \hline
   WHICH? Trial\cite{vb2015}&Heart Failure&18+&280&13\%&\Checkmark&\XSolid&\Checkmark&\Checkmark\\
   \hline
   UCI Machine Learning Repository"Diabetes-130US"\cite{kh2016paper37}&Diabetes&All patients&NA&69,984&\Checkmark&\XSolid&\Checkmark&\Checkmark\\
   \hline
   \hline
    \end{tabular}
    \label{table8}
\end{table*}

\subsection{Specialized Readmission Datasets}
Specialized readmission datasets focus on certain type of diseases, such as diabetes or heart disease, special medical conditions, or a certain type of medical procedures, such as surgery procedure intervention. 

The National Surgical Quality Improvement Program (NSQIP) is a surgical result database of the American College of Surgeons (ACS), which aims to measure the results of risk-adjusted surgical interventions in order to compare results between hospitals. It is based on 135 variables collected before surgery to 30 days after surgery such as demographics, surgical profile and infection at the surgical site. Through this risk adjustment completed by logistic regression model, the results of hospitals of different sizes that serve different patient populations can be fairly compared. ACS NSQIP data enhances the hospital’s ability to zero preventable complications. Because it was developed by surgeons understanding the actual conditions in the operating room, ACS NSQIP can help hundreds of hospitals across the country evaluate the quality of their surgical plans with unparalleled accuracy and significantly improve the surgical results.

\subsection{Data Fields Related to Readmission}
\subsubsection{Dataset Statistics}
Table \ref{table8} includes simple statistics and brief features of the dataset. The type of patient indicates whether the dataset is collected from all causes or from specific cohorts. Population specifies the patient groups, and number of patients/admissions specify the dataset size. Because some patients may return to hospital (readmission) for multiple times, the number of patients and number of readmission often vary for each dataset.

\subsubsection{Patient Information}
The check boxes in Table \ref{table8} list whether a dataset has patient information includes demographics (Demo), patient clinical information (CL) and healthcare utilization (HC). Patients age, gender, insurance information, marital status, education, medical history, income etc, are considered as demographics. Clinical information including patients' in-hospital information such as the treatment he/she received, lab values are categorized into patient information. The last part of patient information is the healthcare utilization, which consists of patients' length of stay in the hospital, admission and discharge date, medical notes etc.

\subsubsection{Hospital Information}
Hospital information (Hosp) specifies whether the dataset has information about the hospitals to which patients were admitted. In most cases, such hospital information includes hospital teaching status, location (rural/urban), hospital ownership (private/public), hospital bed sizes, and the annually hospital total discharges \textit{etc}.


\section{Discussion}
\subsection{Summary of Data \& Model Challenge Solutions}
Data and model challenges represent the two most critical factors for predictive modeling of hospital readmission. In Table \ref{tab:methodsummary}, we outline strength and weakness of solutions proposed to address these two types of challenges. Indeed, no single approach can perfectly address both challenges, and an effective solution should be customized based on the real-world domain settings.
\begin{table*}[h]
    \centering
    \caption{Summary of strength \text{vs.} weakness of methods addressing data \& model challenges}
    \scriptsize
    \begin{tabular}{p{0.1\textwidth}|p{0.22\textwidth}|p{0.32\textwidth}|p{0.3\textwidth}}
    \hline
    \hline
    Challenges&Solutions& Strengths&Weaknesses\\
    \hline
    \multirow{11}{*}{Data challenges}&Sampling approaches&low cost, high understandability&low representativeness\\
    \cline{2-4}
    &Cost-sensitive learning&can integrate medical costs in modeling&time-consuming finding accurate loss function\\
    \cline{2-4}
    &Ensemble learning&better performance than single classifier&time-consuming with large medical record\\
    \cline{2-4}
    &Federated learning&better patient privacy protection&expensive communication\\
    \cline{2-4}
    &Singular features&\multirow{2}{*}{high representative features can be used}& \multirow{2}{*}{sparse, may ignore some informative features}\\
    \cline{2-2}
    &Hybrid features&&\\
    \cline{2-4}
    &Embedding features&dense features with low dimensionality&hard to understand feature meaning\\
    \cline{2-4}
    &Disease specific prediction&\multirow{3}{*}{provide better predictions for targeted patients}&\multirow{3}{*}{not applicable to different disease types prediction}\\
    \cline{2-2}
    &Gender-specific prediction&&\\
    \cline{2-2}
    &Race, ethnicity-specific prediction&&\\
    \cline{2-4}
    &All-cause prediction&rich data from all disease types&less accurate than domain-specific models\\
    \hline
    \multirow{6}{*}{Model challenges}&Simple transparent models&simple, easy to understand&less accurate\\
    \cline{2-4}
    &Hybrid transparent models&high interpretability&large computational requirement\\
    \cline{2-4}
    &Feature selection&\multirow{2}{*}{important, indicative clinical features can be selected}&\multirow{2}{*}{do not guarantee improved performance}\\
    \cline{2-2}
    &Feature learning&&\\
    \cline{2-4}
    &Feature regularization models&\multirow{2}{*}{fair weight values distribution}&\multirow{2}{*}{can not select groups of correlated features}\\
    \cline{2-2}
    &Feature topic models&&\\
    \hline
    \hline
    \end{tabular}
    \label{tab:methodsummary}
\end{table*}
\subsection{Future Research}
This survey opens many opportunities for future study. First, our taxonomy organizes challenges into two main categories (data challenges \textit{vs.} model challenges). Health and medical domains typically rely on specific measures, in addition to common measures, such as accuracy, AUC value etc. Therefore, other challenges, such as performance metrics, may also be considered for future research. Second, our survey emphasizes on academic publications, yet many commercial systems are available and they are focused on the system and engineering aspects of the problem. Third, we are focused on the English literature, and inherently overlooked high-quality publications from non-English venues. 


\section{Conclusions}
In this paper, we provided a comprehensive review of predictive models for hospital readmission. The survey first proposes a taxonomy to summarize challenges into four main categories: (1) data imbalance, locality and privacy; (2) data variety and complexity; (3) model interpretability; and (4) model implementation. We further organized challenges into subgroups in which the main problems of these challenges, the hindrance to hospital readmission prediction research, and improvement or solutions are discussed. Popular predictive model types, according to methods used, and public available datasets for hospital readmission prediction are summarized to provide materials for creative modeling approaches. The survey, including summary and analysis, not only provides a thorough understanding of existing challenges and methods in this field, but also lists available resources to advance the research for accurate hospital readmission prediction and modelling.
\bibliographystyle{IEEEtran}
\footnotesize{\bibliography{reference}}
\end{document}